\documentclass[twosides]{article}
\usepackage[accepted]{aistats2024}
\usepackage[bibliography=common]{apxproof}
\usepackage[T1]{fontenc}
\usepackage{subcaption}
\usepackage{graphicx} 
\usepackage{amsmath,amsfonts,amssymb,amsthm}
\usepackage{dsfont}
\usepackage{float}
\usepackage{babel}
\usepackage[finalizecache=false,frozencache=true,cachedir=minted-cache]{minted}
\usepackage{afterpage}
\usepackage{natbib}
\usepackage{xparse}
\usepackage{adjustbox}
\usepackage{float}
\usepackage{mathtools}
\usepackage{algorithm}
\usepackage{algpseudocode}
\usepackage{booktabs}
\usepackage{xcolor}
\usepackage{tikz}
\usepackage{pgfplots}
\usepackage{comment}
\usepackage{multirow}
\usepackage{siunitx}
\usepackage{hyperref}
\renewcommand{\appendixsectionformat}[2]{PROOFS FOR SECTION~#1\ (#2)}
\usetikzlibrary{positioning, arrows, spy, calc}
\tikzset{boximg/.style={remember picture,red,thick,draw,inner sep=0pt,outer sep=0pt}}
\NewDocumentCommand{\traj}{ O{0} O{T} O{s} O{a} }{ {#3}_{#1},\allowbreak {#4}_{#1},\allowbreak\ldots,\allowbreak {#3}_{#2-1},\allowbreak {#4}_{#2-1},\allowbreak {#3}_{#2}}

\theoremstyle{plain}
\newtheoremrep{theorem}{Theorem}
\newtheoremrep{proposition}{Proposition}
\newtheoremrep{lemma}{Lemma}
\newtheoremrep{corollary}{Corollary}

\theoremstyle{definition}
\newtheoremrep{definition}{Definition}
\newtheoremrep{assumption}{Assumption}
\theoremstyle{remark}
\newtheoremrep{remark}{Remark}
\newtheoremrep{example}{Example}

\newcommand{\mc}[1]{\mathcal{#1}}
\newcommand{\mb}[1]{\mathbb{#1}}
\newcommand{\md}[1]{\mathds{#1}}

\newcommand{\paren}{\text{Parent}}

\newcommand{\term}{\mc T}
\newcommand{\trans}{T}

\newcommand{\unnormprob}{\tilde{p}}
\newcommand{\normprob}{p}
\begin{document}
\twocolumn[

\aistatstitle{Maximum entropy GFlowNets with soft Q-learning}

\aistatsauthor{ \hspace{1em}Sobhan Mohammadpour${^1}{^2}$ \And Emmanuel Bengio${^3}$ \Andf  Emma Frejinger${^4}$ \Andf Pierre-Luc Bacon${^2}{^4}$ }
\runningauthor{Sobhan Mohammadpour, Emmanuel Bengio,  Emma Frejinger, Pierre-Luc Bacon }
\vspace{1.3em}
\aistatsaddress{
\hspace{1em}${^1}$ MIT \And
\hspace{-3em}${^2}$ Mila - Quebec AI Institute  \And
${^3}$ Valence Labs  \AND
{\vspace{-5em}${^4}$ Department of Computer Science and Operations Research, University of Montreal }
}
]
    
\begin{abstract}

Generative Flow Networks (GFNs) have emerged as a powerful tool for sampling discrete objects from unnormalized distributions, offering a scalable alternative to Markov Chain Monte Carlo (MCMC) methods. While GFNs draw inspiration from maximum entropy reinforcement learning (RL), the connection between the two has largely been unclear and seemingly applicable only in specific cases. This paper addresses the connection by constructing an appropriate reward function, thereby establishing an exact relationship between GFNs and maximum entropy RL. This construction allows us to introduce maximum entropy GFNs, which, in contrast to GFNs with uniform backward policy, achieve the maximum entropy attainable by GFNs without constraints on the state space.

\end{abstract}

\section{INTRODUCTION}

Generative Flow Networks (GFNs) have recently emerged as a scalable method for sampling discrete objects from high-dimensional unnormalized distributions. They transform the complexity of navigating such spaces into a sequential decision-making problem, wherein each sequence of actions yields a unique object. Although the notion of framing "inference" as a control problem has been previously explored \citep{fleming1981optimal, buesing2020approximate}, \cite{bengio2021flow} observe that a naive approach based on soft Q-learning \citep[SQL;][]{haarnoja2017reinforcement} and maximum entropy reinforcement learning (RL) tends to favor large objects disproportionately. To counter this, they introduced GFNs, a novel RL-inspired method.

While \cite{bengio2021flow} established the exact equivalence between GFNs and SQL for tree-structured problems, GFNs have primarily been explored outside the theoretical confines of RL. This is due to the ambiguity over how this connection could be applicable more generally. Our work aims to bridge this knowledge gap by developing a valid GFN-like method purely from an RL standpoint. The crucial insight is that the sampling bias, identified by \cite{bengio2021flow}, can be mitigated by formulating a suitable reward function. Combined with the smooth Bellman equation, this leads to samples from the target distribution.
We illustrate that this reward function can be efficiently obtained as the solution to an auxiliary dynamic programming problem, the number of paths leading to a specific node. Our resulting approaches, termed generative SQL, can further be interpreted as an instantiation of the concept of general value functions \citep{sutton2011horde, white2015developing}.

By leveraging our newly forged theoretical connection, we find the backward policy of a GFN whose forward policy corresponds to generative SQL, which we refer to as the maximum entropy backward and can be calculated using the same dynamic program used for generative SQL. We prove that maximum entropy GFNs, GFNs that use the maximum entropy backward, indeed achieve the upper bound of entropy possible for a GFN—a claim hinted at in prior research under stringent conditions  \citep{zhang2022generative}.

Additionally, we reveal that applying Path Consistent Learning \citep[PCL;][]{nachum2017bridging} on our proposed reward yields ``trajectory-balance'' for maximum entropy GFNs.

The principal contributions of this paper are as follows:

\begin{enumerate}
    \item We propose a reward function with the property that, once incorporated within the smooth Bellman equations, leads to policies capable of sampling from the given target distribution.
    \item We provide a formulation backward policy for GFNs with the same policy as the solution of the smooth Bellman equations.
    \item We show that GFNs constructed in this manner have a unique solution- unlike traditional GFNs- and provably reach the maximum entropy in the general case.
    \item We demonstrate through experiments that maximum entropy GFNs enhance the exploration of intermediate states and achieve better results in a hard graph-building environment.
\end{enumerate}

\section{BACKGROUND AND NOTATION}

We provide a list of notation in Table~\ref{tab:notation} and a list of abbreviations in Table~\ref{tab:abrev} in the appendix. 

A Markov decision process (MDP) is a tuple $(\mc S, \mc S_0, \term, \mc A, \md A, \trans, \mb P_0)$ where 
$\mc S$ is the set of states, 
$\mc S_0 \subset \mc S$ is the set of initial states, and
$\term \subset \mc S$ is the set of terminal states. Furthermore, 
$\mc A$ is the set of actions, 
the action mask $\md A: \mc S \rightarrow P(\mc A)$, where $P$ is the power set function, is a function that defines the set of actions available at the state $s$, and
$\trans: \mc S \times \mc A \rightarrow \mc S$ is the deterministic transition function that defines the next state given the current state and actions. 
The initial state of the trajectories is sampled from $\mb P_0:\Delta(\mc S_0)$ where $\Delta(\mc \mc S_0)$ denotes the set of all distributions over the set $\mc S_0$.  We deviate the notation from that of \cite{sutton2018reinforcement} or \cite{bengio2021gflownet} to both find a notation that helps us bridge the two topics and, at the same time, reflect the underlying assumptions and implementations better.

In the context of GFNs, \cite{bengio2021flow} assume that there is a unique initial state $\mc{S}_0=\{s_0\}.$ Also central to the definition of GFNs is the parent function $\paren: \mc S \rightarrow P(\mc S \times \mc A)$. It returns the set of state action pairs $(s,a)$ that reach a state $s'$. It can be thought of as the generalized inverse of $T$ and is defined as 
\[
    \paren(s') = \{(s,a) \in \mc S \times \mc A| a \in \md A(s)\wedge\trans(s,a)=s'\}.
\]
Furthermore, GFNs assume that the transition function $\trans$ is acyclic \citep{bengio2021flow}, which means it is impossible to reach a state from itself. We refer to MDPs with acyclic transition functions as acyclic MDPs for brevity. An essential consequence of the acyclicity assumption is that any regularized dynamic program (DP) converges and has a unique solution \citep{mensch2018differentiable}.

For acyclic MDPs, we define the marginal state distribution $\mu_\pi:\mc S\rightarrow\mb R$ as the probability of passing through a state using the policy $\pi$. The marginal $\mu_\pi(s)$ can be calculated with dynamic programming for deterministic acyclic MDPs using the following recursion:
\[
\mu_\pi(s') = \sum_{\mathclap{(s,a)\in\paren(s')}} \mu(s)\pi(a|s).
\]
We note that $\mu(s) = \mb P_0(s)$ for all $s \in \mc S_0$ and $\sum_{s\in\term}\mu_\pi(s)=1$. 
We call a state  ``reachable'' if there exists a sequence of actions leads to that state from a state in $\mc S_0$ whose initial probability $\mb P_0$ is greater than zero. If this property holds globally, we say that MDP is reachable.

Given an unnormalized distribution on the terminal states $\unnormprob:\Delta(\term)$ called the target, our goal is to find a policy  $\pi: \mc S \rightarrow \Delta(\mc A)$ such that the probability of reaching a terminal state $t\in\term$ is proportional to $\unnormprob(t)$. For simplicity, we assume that $\unnormprob$ is zero for all non-terminal states. We denote the normalized version of $\unnormprob$ as $\normprob$ and write the normalizing constant of the target or the partition function as $Z$.

\subsection{GENERATIVE FLOW NETWORKS}

GFNs enforce that for all terminal states $t\in\term$, the probability of ending the trajectory in $t$, i.e., $\mu(t)$, is proportional to $\unnormprob$. In other words, the policy samples in proportion to $\unnormprob$. They are built around the idea that the MDP is acyclic, i.e., no state can reach itself and that only one initial state $s_0$ exists. Concretely, any Markovian policy $\pi$ that samples terminal states in proportion to $\unnormprob$ fits the detailed balance (DB) constraints
\begin{equation}\label{eq:detailedbalance}\tag{DB}
    F(s)\pi(a|s) = q(s,a|s')F(s'),
\end{equation}
for all transition triplets $s' = T(s, a)$, and where $q:\mc S\rightarrow\Delta(\mc S \times \mc A)$ is the backward policy and $F:\mc S \rightarrow \mb{R}_{\geq0}$ is the state flow function and is assumed to be equal to $\unnormprob(t)$ for all terminal states $t\in\term$. The state flow and marginal are linked $F(s)/F(s_0)=\mu(s)$. Any $\pi$ and $\unnormprob$ uniquely identify the backward policy and the state flow function. Furthermore, for any $\pi$, $q$, and $F$ that fit (\ref{eq:detailedbalance}), $\pi$ samples in proportion to $\unnormprob$ \citep{bengio2021gflownet}. 

Detailed balance (\ref{eq:detailedbalance}) is not the only formulation possible for a GFN; Trajectory Balance  \citep[\ref{eq:trajectorybalance};][]{malkin2022trajectory} is obtained by multiplying the (\ref{eq:detailedbalance}) constraint over a trajectory i.e.
\begin{equation}\label{eq:trajectorybalance}\tag{TB}
    Z\prod_{t=0}^{T-1} \pi(a_t|s_t)=\unnormprob(s_T)\prod_{t=0}^{T-1}q(s_t,a_t|s_{t+1}),
\end{equation}
where $Z$ is shorthand for $F(s_0)$. Additional formulations for GFNs, including \citet{bengio2021flow}'s original constraint, are included in Appendix~\ref{appen:gfn}. 

\citet{bengio2021flow} propose minimizing the residual of a GFN constraint like (\ref{eq:detailedbalance}), which can admit an infinite number of solutions (see Appendix~\ref{appen:gfn}). On the other hand, for any strictly concave function, there is only a unique GFN that maximizes that function. One example of a strictly concave function is the flow entropy, defined as
\[
\mb H(\pi)=\mb E\left[ \sum_{t=0}^{T-1} H(\pi(\cdot|s_t)) \right] = \sum_{s\in\mc S} \mu_\pi(s)H(\pi(\cdot|s)),
\]
where $H(\pi)= -\sum_{a\in\md A(s)} \pi(a|s)\log \pi(a|s)$ is the entropy. In Section~\ref{sec:fe}, we not only show that flow entropy is strictly concave, we show it is equal to the entropy of the policy. Flow entropy was first introduced by \citet{zhang2022generative} where they showed that for a restrictive class of MDPs setting the backward to be uniform on the parents (i.e., $q(s, a|s') = 1 / |\paren(s')|$) maximizes the flow entropy. However, \citet{zhang2022generative}
restrict the class of MDPs they analyze so much as to exclude the drug design MDP of \citet{bengio2021flow}. 
Concretely, \citet{zhang2022generative} look at MDPs where the state is a vector of values and placeholders, and each action sets a placeholder to a value. For instance, the state could be $(\star, \star, 0)$, for the placeholder $\star$, and the action ``set element $0$ to $1$'' would yield the state $(1, \star, 0)$. While many MDPs used on GFNs add elements to placeholders, this definition is much more restrictive. \citet{zhang2022generative}'s construction can be relaxed to having MDPs with finite and fixed horizons where the transition function is layered, meaning that there exists a partitioning of the state space $\mathcal{L}_0,\mathcal{L}_1,\mathcal{L}_2,\ldots,\mathcal{L}_n$ where $\mathcal{L}_0=\mc S_0$, $\mc L_n=\term$ and the parents of any set in layer $\mathcal{L}_i$ is in $\mathcal{L}_{i-1}$ for $i \geq 1$. Furthermore, they assume that the number of actions in layer $\mathcal{L}_i$ is $n - i$. These assumptions are restrictive; for instance, adding the constraint that the number of ones is always less than the number of zeros invalidates the assumptions.

\subsection{SOFT Q-LEARNING}

Given a reward function over transitions $R:\mc S \times \mc A \rightarrow \mb R$, terminal reward $R_{\mc T}: \mc T \rightarrow \mb R$, along with the discount factor $\gamma\in(0,1]$, we can derive two fundamental functions: the state value function $V(s)$, and the state-action value function $Q(s,a)$. The state value function $V(s)$ represents the maximum discounted reward obtainable from the state $s$, while the state-action value function $Q(s,a)$ represents the maximum discounted reward from state $s$ when taking the action $a$. The Bellman equation \citep{bellman1954theory} articulates the relationship between the $V$ and $Q$ functions. This equation can be formulated as:
\begin{subequations}\label{eqs:bellman}
\begin{align}
    Q(s,a) &= R(s, a) + \gamma V(T(s,a)), \\
    V(s) &= \max_{a\in\md A(s)} Q(s, a).\label{eq:bellman:v}
\end{align}
\end{subequations}
We assume that the value of the V function is the terminal reward at the terminal states, i.e., $V(t)=R_{\mc T}(t)$ for all $t\in\term$.

If we augment the reward with the entropy or an i.i.d.\ Gumbel term to the maximum in (\ref{eq:bellman:v}), we obtain the soft (smooth) Bellman equation \citep{rust1987optimal,todorov2006linearly,ziebart2008maximum,peters2010relative,rawlik2012stochastic,fosgerau2013link,van2015learning,fox2015taming,nachum2017bridging,haarnoja2017reinforcement,geist2019theory,garg2023extreme}
\begin{subequations}\label{eq:softbellman}
\begin{align}
    Q(s,a) &= R(s, a) + \gamma V(T(s,a))  \\
    V(s) &= \max_{\pi \in\Delta(\md A(s))} \mb E_{a\sim\pi}[Q(s, a) - \tau\log\pi_a], 
\end{align}
\end{subequations}
at temperature $\tau$. If $\gamma$ is less than one, the soft Bellman equation will have a unique solution \citep{geist2019theory}. However, the existence of a unique solution is not guaranteed in the undiscounted case, i.e., where $\gamma=1$. \citet[Remark 2]{mai2022undiscounted} give a sufficient condition for the existence of a unique solution. In the acyclic case, \cite{mensch2018differentiable} showed that the soft Bellman equation has a unique solution. We note that there cannot be more than one solution to the equations as entropy is strictly concave.

Given a Q function, the value function $V$ is equal to $V(s)=\tau\log\sum_{a\in\md A(s)}\exp(Q(s,a)/\tau),$
and the policy is $\pi(a|s)=\frac{\exp(Q(s,a)/\tau)}{\sum_{a'\in\md A(s)} \exp\left(Q(s,a')/\tau\right)}=\exp\left(Q(s,a)/\tau - V(s)/\tau\right).$ Taking the log on both sides of this expression is the basis for path consistency learning (PCL) \citep{nachum2017bridging}, which aims to enforce a temporal consistency between the policy and value function over multiple steps. For deterministic MDPs, path consistency learning enforces %
\begin{multline}\label{eq:pcl}\tag{PCL}
V(s_i) + \sum_{t=i}^{j-1} \tau\gamma^{t-i}(\tau\log\pi(a_t|s_t) - R(s_t, a_t)) \\ = \gamma^{j-i} V(s_j)
\end{multline}
instead of the soft Bellman equation over a sub trajectory $\traj[i][j]$. We note that the relationship between (\ref{eq:trajectorybalance}) and (\ref{eq:detailedbalance}) is reminiscent of (\ref{eq:pcl}) and (\ref{eq:softbellman}).

In the remainder of this paper, we assume $\tau=1$ and $\gamma=1$ for clarity.

A key property we use in our proofs is that the probability of a trajectory can be determined using the value function of its starting and ending states, along with the rewards accrued throughout that trajectory. This is further demonstrated in the following proposition.
\begin{propositionrep}\label{prop:fosgerau}
\citep{fosgerau2013link} in deterministic entropy regularized MDPs with no discounting we have $\log \mb P(\traj[i][j]) = \sum_{t=i}^{j-1} \log \pi(a_t|s_t) = V(s_j) + \sum_{t=i}^{j-1}R(s_t,a_t) - V(s_i)$.
\end{propositionrep}
\begin{proof}
 This can be verified easily as
\begin{align}
    \log \mb P(\traj[i][j]) 
    &= \sum_{t=i}^{j-1} \log\pi(a_t|s_t) \\
    &= \sum_{t=i}^{j-1} Q(s_t,a_t) - V(s_t)\\
    &= \sum_{t=i}^{j-1} R(s_t, a_t) + V(s_{t+1}) - V(s_t) \label{eq:q:something}
\end{align}
simplifying (\ref{eq:q:something}) yields
\begin{equation}\label{eq:sqltrajectoryprob}
    \log \mb P(\traj[i][j]) = V(s_j) + \sum_{t=i}^{j-1} R(s_t,a_t) - V(s_i).
\end{equation}

Note that for whole trajectories $r=\traj$, (\ref{eq:sqltrajectoryprob}) is equal to 
\begin{equation}
    \log \mb P(\traj) = R_\term(S_T) + \sum_{t=0}^{T-1} R(s_t,a_t) - V(s_0).
\end{equation}
\end{proof}

By Proposition~\ref{prop:fosgerau}, the likelihood of trajectories that share initial and terminal states is proportional to the exponent of their reward difference. If there are no intermediate rewards, they have the same probability. This property is important for our proofs.

\section{FROM SOFT Q-LEARNING TO MAXIMUM ENTROPY GFNs}
\label{sec:framework}

In this section, we derive a policy that reaches terminal states in proportion to an unnormalized distribution $\unnormprob$ by constructing an appropriate reward under the soft Bellman equations. As a reminder, we assume an acyclic deterministic MDP with only one initial state. The acyclicity assumption reflects that we are building complex discrete objects by accumulation of primitive parts. As for the assumption of a single initial state, itc reflects the fact that we start over every time. These assumptions mirror the assumptions of GFNs. All proofs are in the appendix.

If we set $R(s,a)=0$ and $R_{\mc T}(s_T)=\unnormprob(s_T)$, the return for a trajectory $\traj$ is $\unnormprob(s_T)$. By Proposition~\ref{prop:fosgerau}, the likelihood of taking the trajectory is proportional to $\exp \unnormprob(s_T)$. Thus, we can estimate the probability of sampling any terminal state, given the number of trajectories that lead to that state.
\begin{propositionrep}
\citep[Proposition~1 in][]{bengio2021flow} For a terminal state $t\in\term$ and the numer of trajectories starting at $s_0$ that terminate in $t$, $n(t)$, the probability of a trajectory terminating in $t$, if we set the terminal reward $R_{\mc T}(t)$ to $\unnormprob(t)$ and have no intermediate reward, is proportional to $n(t)\exp(\unnormprob(t))$.
\end{propositionrep}\label{prop:sqlp}
\begin{proof}
    The probability of taking any of the trajectories ending in $t$ is proportional to $\exp(\unnormprob(t))$, and there are $n(t)$ trajectories; thus, the probability of reaching the state $t$ is proportional to $n(t)\exp(\unnormprob(t))$. 
\end{proof}

We include the proof since it is more concise than the one in \cite{bengio2021flow}. In the next proposition, we modify our reward function such that SQL samples terminal states in proportion to $\unnormprob$.

\begin{propositionrep}
For terminal reward $R_{\mc T}(t)=\log \unnormprob(t) - \log n(t)$, the probability of reaching the state $t$ is proportional to $\unnormprob(t)$.   
\end{propositionrep}
\begin{proof}
    If $R_{\mc T}(t)=\log\unnormprob(t) - \log n(t)$, then the probability of any trajectory ending in $t$ is proportional to $\unnormprob(t)/n(t)$, and thus the probability of reaching $t$ is $\unnormprob(t)$.
\end{proof}

\begin{definition}
    We call soft Q-Learning with the rewards $R(s,a)=0$ and terminal rewards $R_{\mc T}(s)=\log\unnormprob(s) - \log n(s)$ generative soft Q-learning or GSQL.
\end{definition}

In Appendix~\ref{appen:n} we show that it is possible to calculate $n(s)$ for certain MDPs with combinatorial structures but in general $n(s)$ can be calculated using DP.
\begin{theoremrep}\label{theo:n}
The number of trajectories $n(s)$ satisfies
\begin{equation}
    n(s) = \sum_{(s',a')\in\paren(s)} n(s'),\label{eq:n:2}
\end{equation}
and $n(s_0)=1$.
\end{theoremrep}
\begin{proof}
Let $\md L(s)$ be the set of paths from $s_0$ to $s$; every trajectory in $\md L(s)$ comes from one of the parents of $s$. Thus, every trajectory in $\md L(s)$ is the concatenation of a trajectory that leads to a parent of $s$, one action that leads to $s$, and $s$. Hence, for every trajectory to a parent of $s$ and an action $a$ that leads to $s$, we have a unique trajectory that leads to $s$.
\end{proof}

For most MDPs, the recursion of Theorem~\ref{theo:n} is not tractable as the size of the DP table grows exponentially with respect to the horizon. Instead, we advocate learning $n$. Using the inverted MDP defined below, we show that we can leverage entropy regularized RL methods to learn $n$.
\begin{definition}
    The inverse of an MDP 
    \[(\mc S, \mc S_0, \term, \mc A, \md A, \trans)\]
    is an MDP 
    \[(\mc S, \term, \mc S_0, \mc S \times \mc A, \paren, \bar{\trans})\]
    where the set of actions is the set of state action pairs of the original MDP, the action mask function is the set of parents, and the inverted dynamics $\bar{\trans}$ undoes the action such that $\bar{\trans}(s', (s, a))=s$. We omit $\mb P_0$ because it does not affect the calculations. 
\end{definition}

Note that the inverse MDP is implicitly present in the definition of GFNs as the backward policy is defined on the inverted MDP and that the inverse of the inverted MDP is the original MDP. Every trajectory $\traj$ in the original MDP has a corresponding trajectory $s_T,(s_{T-1},a_{T-1}),s_{T-1},\ldots,(s_0, a_0),s_0$ in the inverted MDP.
Using the inverted MDP, the dependence of $n(s)$ on the parents becomes a dependence on the children, giving rise to a formulation that uses a Bellman equation.

It is convenient to learn $\log n$ both for numerical purposes and for the synergy it has with Soft Q-learning. In the log space, the sum in~(\ref{eq:n:2}) is replaced by a log-sum-exp. Indeed, as the following proposition shows, $\log n$, denoted as $l$, fits the soft Bellman equation.
\begin{propositionrep}\label{prop:nrevsql}
    Let $l(s)=\log n(s)$, then $l$ is the value function of the soft Bellman equation in the inverted MDP with the rewards set to zero for all states and transitions.
\end{propositionrep}
\begin{proof}
    Since $l$ is the value function of the soft Bellman equation,
    \begin{equation}
        l(s') = \log \sum_{(s,a)\in \paren(s')} \exp (l(\bar{\trans}(s', (s,a)))) = \log \sum_{(s,a)\in \paren(s')} \exp l(s),
    \end{equation}
    holds since the reward is zero. Taking the exponent of both sides yields
    \begin{equation}
         n(s') = \exp l(s') = \sum_{(s, a) \in\paren(s')} \exp l(s) = \sum_{(s, a) \in\paren(s')} n(s).
    \end{equation}
    Furthermore, since the value of the terminal state is zero, the value of the original initial states is zero, i.e., $\exp l(s_0)= \exp 0 = 1 = n(s_0).$ A unique solution is guaranteed by Propostion~2 of \citet{mensch2018differentiable}.
\end{proof}
\begin{figure*}
    \centering
    
    \centering
        \begin{subfigure}[t]{0.3\textwidth}
    \centering
	\begin{tikzpicture}[->,>=stealth',shorten >=1pt,node distance=4cm,
		thick,main node/.style={circle,fill=blue!20,draw,minimum size=1cm,inner sep=0pt},scale=0.6]
		 \node[main node] (1) at (0,0) {$s_0$};
		\node[main node] (2) at (3,2) {$s_1$};
		\node[main node] (3) at (3,-2) {$s_2$};
		\node[main node] (4) at (6,0) {$s_T$};
		
		\path[every node/.style={font=\sffamily\small,fill=white, sloped}]
		(1) edge node  {$1/4$} (2)
		(1) edge node  {$3/4$} (3)
		(3) edge node  {$2/3$} (4)
		(3) edge node  {$1/3$} (2);
    \path[every node/.style={font=\sffamily\small, sloped}]
        (2) edge node {} (4);
	\end{tikzpicture}
 \end{subfigure}%
 \hfill%
 \begin{subfigure}[t]{0.3\textwidth}
	\begin{tikzpicture}[->,>=stealth',shorten >=1pt,node distance=4cm,
		thick,main node/.style={circle,fill=blue!20,draw,minimum size=1cm,inner sep=0pt},scale=0.6]
		 \node[main node] (1) at (0,0) {$s_0$};
		\node[main node] (2) at (3,2) {$s_1$};
		\node[main node] (3) at (3,-2) {$s_2$};
		\node[main node] (4) at (6,0) {$s_T$};

		\path[every node/.style={font=\sffamily\small}]
		(1) edge node {} (2)
		(1) edge node {} (3)
		(2) edge node {} (4)
		(3) edge node {} (4)
		(3) edge node {} (2);
	\end{tikzpicture}
 \end{subfigure}%
\hfill%
    \begin{subfigure}[t]{0.3\textwidth}
    \centering
	\begin{tikzpicture}[->,>=stealth',shorten >=1pt,node distance=4cm,
		thick,main node/.style={circle,fill=blue!20,draw,minimum size=1cm,inner sep=0pt},scale=0.6]
		 \node[main node] (1) at (0,0) {$s_0$};
		\node[main node] (2) at (3,2) {$s_1$};
		\node[main node] (3) at (3,-2) {$s_2$};
		\node[main node] (4) at (6,0) {$s_T$};

		\path[every node/.style={font=\sffamily\small,fill=white,sloped}]
		(1) edge node  {$1/3$} (2)
		(1) edge node {$2/3$} (3)
		(3) edge node {$1/2$} (4)
		(3) edge node  {$1/2$} (2);
        \path[every node/.style={font=\sffamily\small}]
        (2) edge node {} (4);
	\end{tikzpicture}
 \end{subfigure}
    \caption{Comparison of maximum entropy and uniform backward. Left: uniform backward policy, middle: MDP, right: maximum entropy gflownet. The numbers are the probabilities of the policies at state $s_0$ and $s_2$.}

    \label{fig:unfi}
\end{figure*}
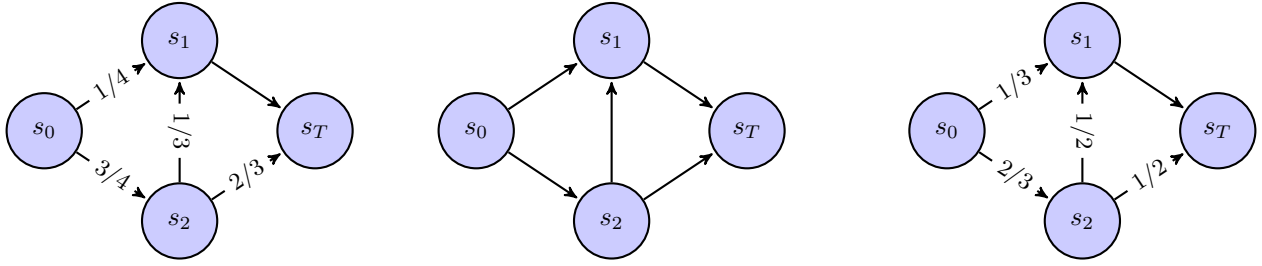

Proposition~\ref{prop:nrevsql} allow us to use standard entropy regularized RL tools to learn  $n$. This allows us to assert that learning $n$ is \textbf{not} harder than learning the GFN. Indeed, as shown in the experiment section, it is often easier. Any constraint like the soft Bellman equation and PCL can be used. If PCL is used, let $\log q((s,a)|s')=l(s) - l(s')$, the PCL equation becomes $l(s_j)+ \sum_{t=i}^{j-1}\log q((s_t,a_t)|s_{t+1}) = l(s_i)$. We revisit this $q$ in Section~\ref{section:bwgsql}. Notice how the value of the initial state in the trajectories is on the right-hand side, not the left-hand side, as we are using a consistency equation on the inverted MDP.

\begin{figure*}[ht]
\begin{center}
    \begin{subfigure}[T]{0.15\textwidth}
    \includegraphics[width=\textwidth]{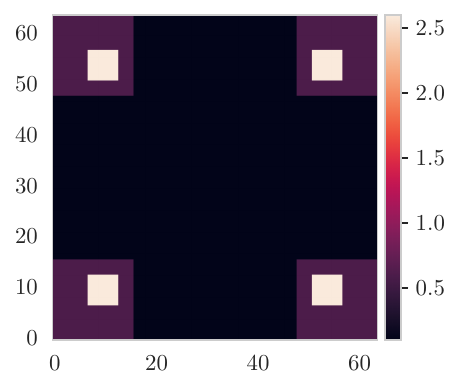}    
    \caption{$\unnormprob$\label{fig:64:p}}    
    \end{subfigure}
    \begin{subfigure}[T]{0.15\textwidth}
    \includegraphics[width=\textwidth]{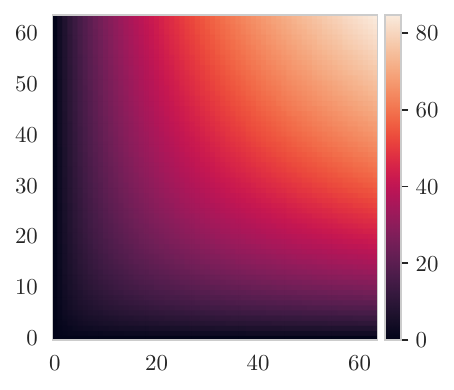}    
    \caption{$l$\label{fig:64:l}}    
    \end{subfigure}
    \begin{subfigure}[T]{0.15\textwidth}
    \includegraphics[width=\textwidth]{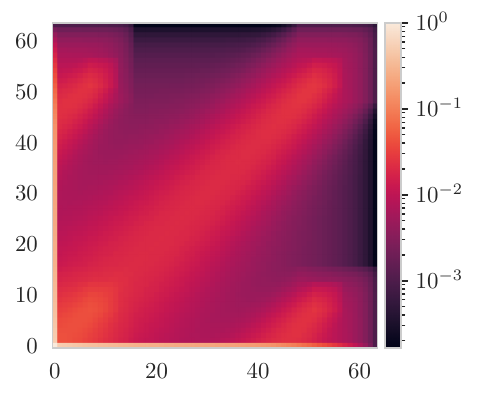}
    \caption{uniform $\mu$\label{fig:64:umu}}      
    \end{subfigure}
    \begin{subfigure}[T]{0.15\textwidth}
    \includegraphics[width=\textwidth]{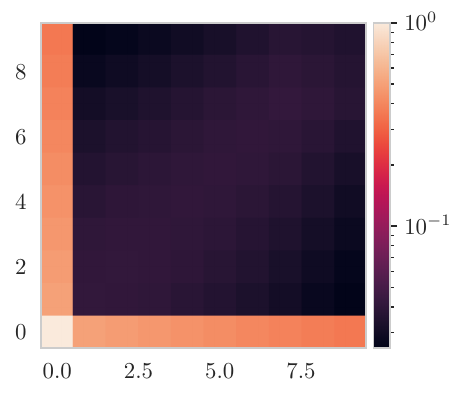}
    \caption{\centering uniform $\mu$ zoomed\label{fig:64:umu:zoom}}   
    \end{subfigure}
    \begin{subfigure}[T]{0.15\textwidth}
    \includegraphics[width=\textwidth]{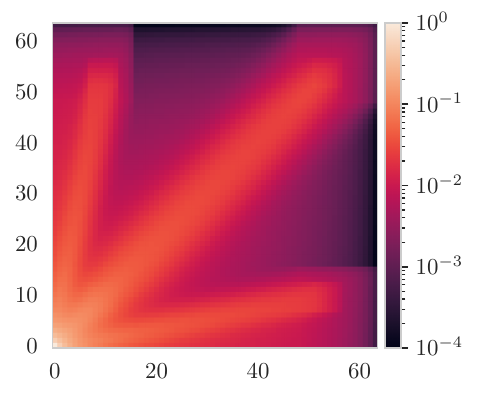}
    \caption{maxent $\mu$\label{fig:64:max}}        
    \end{subfigure}
    \begin{subfigure}[T]{0.15\textwidth}
    \includegraphics[width=\textwidth]{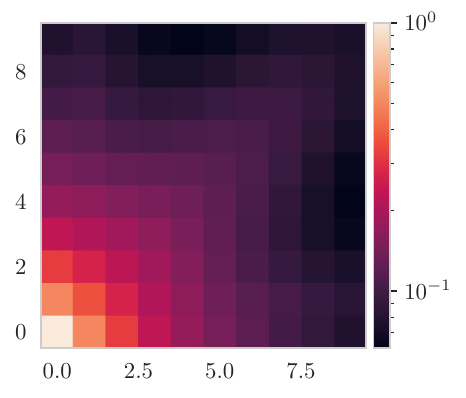}
    \caption{\centering maxent $\mu$ zoomed\label{fig:64:max:zoom}}   
    \end{subfigure}

\end{center}
    \caption{From left to right, target, $l=$, and marginal of the uniform backward and maximum entropy GFNs for the $64^2$ grid. Note the log scale colors for $\mu$ and the non-smooth partitioning of the flow around the bottom and left edges with the uniform backward policy.\label{fig:64}}

    \begin{center}
    \begin{subfigure}[T]{\textwidth}
    \centering
    \includegraphics[width=0.5\textwidth]{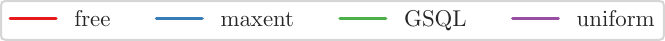}
    \end{subfigure}
    \begin{subfigure}[T]{0.24\textwidth}
    \centering
    \includegraphics[width=\textwidth]{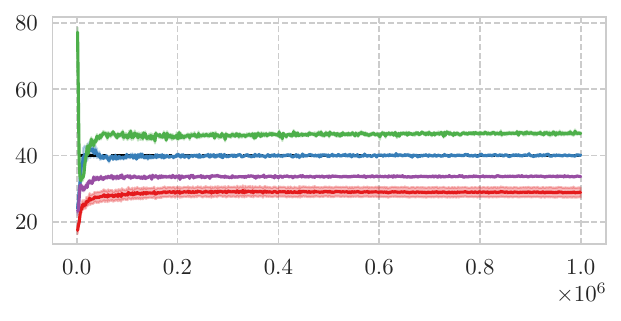}
    \end{subfigure}
    \begin{subfigure}[T]{0.24\textwidth}
    \centering
    \includegraphics[width=\textwidth]{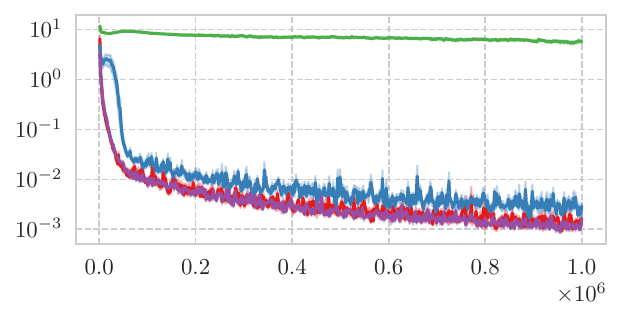}
    \end{subfigure}
    \begin{subfigure}[T]{0.24\textwidth}
    \centering
    \includegraphics[width=\textwidth]{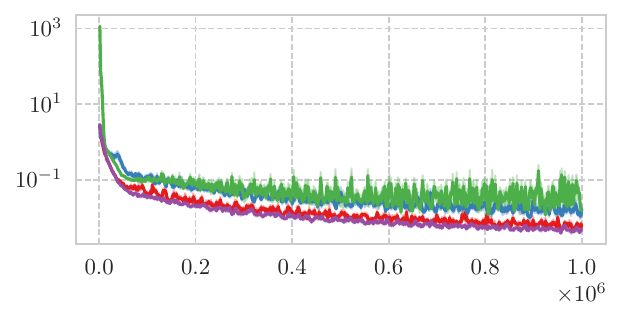}
    \end{subfigure}
    \begin{subfigure}[T]{0.24\textwidth}
    \centering
    \includegraphics[width=\textwidth]{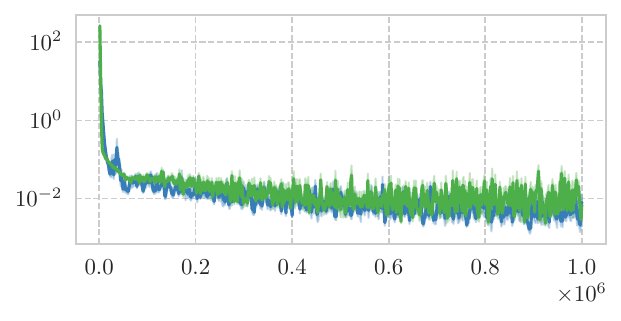}
    \end{subfigure}
    \begin{subfigure}[T]{0.24\textwidth}
    \centering
    \includegraphics[width=\textwidth]{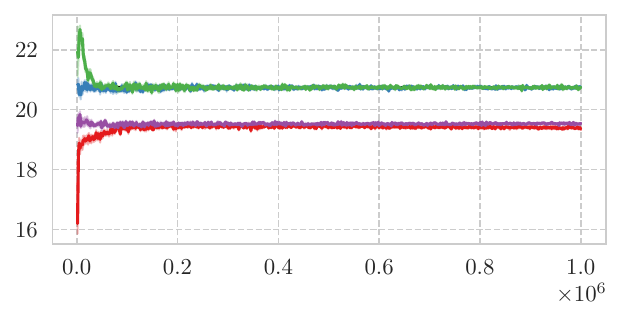}
    \caption{entropy}
    \end{subfigure}
    \begin{subfigure}[T]{0.24\textwidth}
    \centering
    \includegraphics[width=\textwidth]{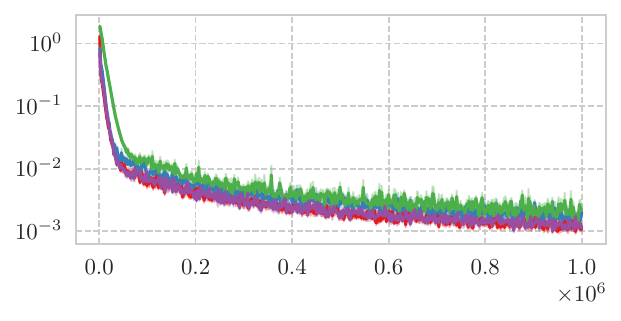}
    \caption{KL}
    \end{subfigure}
    \begin{subfigure}[T]{0.24\textwidth}
    \centering
    \includegraphics[width=\textwidth]{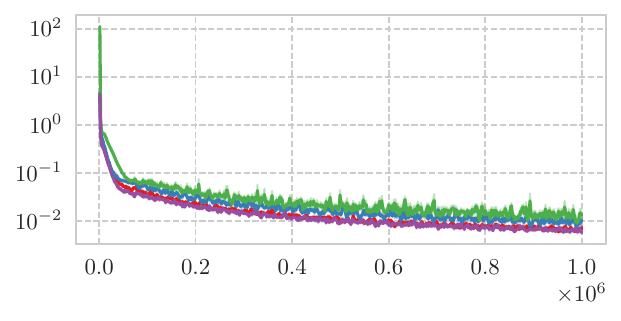}
    \caption{policy loss}
    \end{subfigure}
    \begin{subfigure}[T]{0.24\textwidth}
    \centering
    \includegraphics[width=\textwidth]{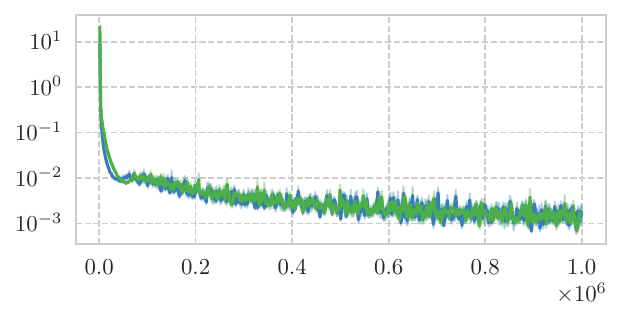}
    \caption{$n$ loss   }
    \end{subfigure}
        \end{center}
    \caption{Various metrics for the hypergrid experiments. The top row is for the $64^2$ grid, and the bottom row is for the $8^4$ grid. The black line shows the maximum entropy attainable by a GFN. Log y-scale was used.}
    \label{fig:enter-label2}
\end{figure*}

The next proposition shows that the value initial state $s_0$ of GSQL equals the logarithm of total flow $\log F(s_0)$ or $\log Z$. We begin with a lemma that shows that the value of every state $s$ is the log-sum-exp of all trajectories that pass through $s$. Using trajectories instead of terminal states allows us to not run into issues when a state is the descendent of multiple states that are not the descent of each other.

\begin{lemmarep}\label{lemma:1}
    In finite acyclic MDPs with $R(s, a)=0$, $V(s)$ is equal to the log-sum-exp of the reward of all trajectories that start at $s$.
\end{lemmarep}
\begin{proof}
    We show the lemma by induction on the depth of the topological sort of the states. The result is trivial for terminal states $t\in\term$ as $V(t)=R_\term(s)$. If the proposition is true for all states with a depth of at least $i$, for any state $s$ at depth $i-1$ we have that $V(s)=\log \sum_{a\in\md A(s)} \exp Q(s,a) = \log \sum_{a\in\md A(s)}\exp V(s')$ where $s'=\trans(s,a)$. Since we know that $V(s')$ is the logarithm of the sum of the rewards of all trajectories that start at $s'$, $\exp V(s')$ is the sum of the reward of all trajectories that start with $s,a$. Since the sum is over all actions, $V(s)$ is the logarithm of the sum of the rewards of all trajectories starting at $s$.
\end{proof}

Equipped with Lemma~\ref{lemma:1}, we can show the following proposition for GSQL.
\begin{propositionrep}\label{prop:VisZ}
The value function of the initial state is equal to the logarithm of the partition function, i.e., $V(s_0) = \log \sum_{s\in\term} \unnormprob(s) =\log Z$.
\end{propositionrep}
\begin{proof}
    For each terminal state $t\in\term$, there are $n(t)$ trajectories, each with reward $\log \unnormprob(t) - \log n(t)$; thus, the sum-exp of their reward is $n(t)\exp(\log \unnormprob(t) - \log n(t))=\unnormprob(t)$. Taking the sum of the sum-exp of reward over the set of all terminal states results in the sum-exp over the set of all trajectories: 
    \begin{align}    
        V(s_0) &= \log \sum_{s_0,a_0,\ldots,s_T} \unnormprob(s_T)/n(s_T) \\
        &= \log \sum_{s_T\in\term}\sum_{s_0,a_0,\ldots,s_T} \unnormprob(s_T)/n(s_T), \\
        &= \log \sum_{t\in\term} \unnormprob(t) = \log Z.
    \end{align}
    The third equality holds because there are $n(s_T)$ trajectories for the destination $s_T$, we remove the second summation and the corresponding division.
\end{proof}

\subsection{A different definition of flow entropy}\label{sec:fe}

This subsection shows that flow entropy equals entropy over trajectories and that GSQL achieves maximum entropy. We define the trajectory entropy as follows:
\begin{definition}
The trajectory entropy $H$ is the entropy over a set of trajectories and is defined as 
\begin{equation}
     H(\pi) = \mb E\left[\sum_{t=0}^{T-1} \log \pi(a_t|s_t)\right]
\end{equation}
for Markovian policies. 
\end{definition}
The definition is trivially extendable to non-Markovian policies as it is the entropy of the distribution over trajectories. The following proposition shows that $\mb H$ and $H$ are, in fact, equal.
\begin{propositionrep}\label{prop:flowentisent}
    The trajectory entropy and flow entropy are equal for Markovian policies, i.e., $\mb H(\pi)=H(\pi)$.
\end{propositionrep}
\begin{proof}
 We start with the definition of flow entropy, let $\pi(\cdot|s_t)$ be the distribution over actions at state $s_t$, and $\bar{\term}$ the complement of $\term$ or the set of all non terminal states, we have:
\begin{align}
    \mb H(\pi) &= \mb E_{s_0,a_0,\ldots,s_T}[\sum_{t=0}^{T-1}H(\pi(\cdot|s_t))].\\
    \shortintertext{Using the linearity of the expectation, we move the summation out of the expectation to get:}
    &= \sum_{s\in\bar{\term}} \mb P(\cdots s\cdots) H(\pi(\cdot|s))  \\
    \shortintertext{Since the probability of all trajectories that pass through $s$ is equal to the probability to partial trajectories that end in $s$, we can further simplify the summation to:}
    &= \sum_{s\in\bar{\term}} \mb P(\cdots s) H(\pi(\cdot|s)\\
    &= \sum_{s\in\bar{\term}} \mb P(\cdots s) \sum_{a\in\md A(s)} \pi(a|s)\log\pi(a|s) \\
    \shortintertext{Using the Bayes' rule, we have that $\mb P(\cdots s)\pi(a|s) = \mb P(\cdots sa)$.}
    &= \sum_{s\in\bar{\term},a\in\md A(s)} \mb P(\cdots sa)\log\pi(a|s) \\
    &= \sum_{s\in\bar{\term},a\in\md A(s)} \mb P(\cdots sa\cdots)\log\pi(a|s) \\
    &= \mb E_{s_0,a_0,\ldots,s_T}[\sum_{t=0}^{T-1}\log\pi(a_t|s_t)] \\
    &= H(\pi).
\end{align}
\end{proof}
\begin{theoremrep}
    GSQL achieves the maximum entropy a policy sampling in proportion to the target can achieve.
\end{theoremrep}
\begin{proof}
        Given the elementary property that $H(X, Y)=H(X)+\mb E[H(Y|X)]$, the trajectory entropy is the sum of entropy of the destination $H(\normprob)$ (recall that $\normprob$ is a distribution proportional to $\unnormprob$) and the expected conditional entropy on the destinations $\mb E[H(\pi|s_T)]$ i.e.
\begin{equation}
    H(\pi) = H(\normprob) + \mb E_{s_T\sim \normprob}[H(\pi|s_T)].
\end{equation}
Since  $H(\normprob)$, and the distribution of $s_T$ is part of the problem, we can maximize the interior of the expectation. The uniform distribution achieves maximum entropy, and our SQL samples uniformly on trajectories conditioned on the destination (as they have the same reward), thus, GSQL  achieves maximum entropy.
\end{proof}

The uniqueness of the maximum entropy GFN is derived from the following concavity proof, which is the same as the proof given by \cite{hoda2010smoothing} except that we define the flow for acyclic graphs, not trees.
\begin{lemmarep}\label{prop:flowregconvex}
    For any concave function $H: \Delta \rightarrow \mb R$, the function $\sum_{s\in\mc S} F(s) H\left(\frac{F(s,\cdot)}{F(s)}\right)$ where $F(s,\cdot)$ is the vector of outgoing flows at state $s$ and $F(s)$ is the sum of all outgoing flows, is concave and maximizing any such function yields a unique GFN.
\end{lemmarep}
\begin{proof}
    The proof uses the dilation or perspective operation \citep[Section~2.2 of][]{hiriart2004fundamentals}. For a strictly concave function $H(x)$, the dilated  function $yH(x/y)$, where $x\in\mb{R}^n_{\geq0}$ and $y \in\mb{R}_{\geq0}$, is also strictly concave. We assume that $yH(x/y)$ is zero when $y$ is zero.
    
    The function $H$ is strictly concave, so the dilated function $F(s)H(F(s,\cdot)/F(s))$ is also strictly concave. 
    
    Since the flow entropy is strictly concave with respect to the state and state-action flow vectors, maximizing the flow entropy under the GFN constraints will yield a unique policy as the GFN constraints are linear.
\end{proof}

Given that the policy entropy is the flow entropy, we can specialize Lemma~\ref{prop:flowregconvex}.
\begin{propositionrep}
Markovian entropy and flow entropy is a special case of Lemma~\ref{prop:flowregconvex} and thus is strictly concave, and thus the maximum entropy GFN is unique.
\end{propositionrep}
\begin{proof}
    Since $\pi(a|s)=F(s,a)/F(s)$, the entropy (see Proposition~\ref{prop:flowentisent}), $\sum_{s\in\mathcal{S}} F(s)H({F(s,\cdot)}/{F(s)})$, is strictly concave.
\end{proof}

\subsection{The backward of GSQL}\label{section:bwgsql}

As mentioned in the introduction, for a fixed policy and $\unnormprob$, we can uniquely identify the backward policy $q$. The following identifies the backward policy of GSQL.
\begin{theoremrep}    
The backward policy $q$ of the policy of GSQL is 
\begin{equation}
    q(s,a|s')=n(s)/n(s').
\end{equation}
\end{theoremrep}
\begin{proof}
By the definition of $n$, $q$ is a distribution as it is both positive and sums to one over the set of all parents. To show that $q$ is the backward policy of GSQL, we start with Proposition~\ref{prop:VisZ} and show that  (\ref{eq:trajectorybalance}) holds for the proposed backward policy.

By Proposition~\ref{prop:VisZ} we have that
\begin{equation}
    Z = \exp V(s_0).
\end{equation}
Multiplying both sides by $\exp\left(\log \unnormprob(s_T) -  V(s_0) - \log n(s_T)\right)$ and simplifying the $\exp \log$ on the left hand side yields
\begin{equation}
    Z\exp(\log \unnormprob(s_T) - \log n(s_T) - V(s_0)) = \unnormprob(s_T)/n(s_T).
\end{equation}
Since $V(s_T)=\log \unnormprob(s_T) - \log n(s_T)$ and $n(s_0)=1$ we get
\begin{equation}
    Z  \exp(V(s_T)-V(s_0)) = \unnormprob(s_T) n(s_0)/n(s_T).
\end{equation}
We then use the fact that $V(s_T) - V(s_0) = \sum_{t=0}^{T-1}V(s_{t+1}) - V(s_t)$ and $n(s_t)/n(s_T)=\prod_{t=0}^{T-1} n(s_t)/n(s_{t+1})$ to get
\begin{equation}
    Z \prod_{t=0}^{T-1} \exp(V(s_{t+1}))/\exp(V(s_t)) = \unnormprob(s_T) \prod_{t=0}^{T-1} n(s_t)/n(s_{t+1}).
\end{equation}
Using the definition of $q$ and Proposition~\ref{prop:fosgerau} we get
\begin{equation}
    Z \prod_{t=0}^{T-1} \pi(a_t|s_t) = \unnormprob(s_T) \prod_{t=0}^{T-1} q(a_t,s_t|s_{t+1}),
\end{equation}    
which is (\ref{eq:trajectorybalance}).
\end{proof}

The backward proposed here can be used to train maximum entropy GFNs.

\begin{remarkrep}
     The backward policy $q$ gives a uniform distribution over all backward trajectories leading to $s_0$, given $s_T\in\term$. 
\end{remarkrep}
\begin{proof}
     Since $n(s_t)/n(s_{t+1})$ of the trajectories to $s_{t+1}$ come through the action $a_t$, the uniform distribution over the trajectories needs to follow that ratio as well.
\end{proof}

\begin{remarkrep}\label{remark:nsub}
    For a sub trajectory $\traj[i][j]$, $\prod_{t=i}^{j-1} q(s_t,a_t|s_{t+1})=n(s_i)/n(s_j)$.
\end{remarkrep}
\begin{proof}
$\prod_{t=i}^{j-1} q(s_t,a_t|s_{t+1}) = \prod_{t=i}^{j-1} n(s_t)/n(s_{t+1})=n(s_i)/n(s_j).$
\end{proof}

\begin{remark}
    We find the uniform backward if the number of paths to all parent nodes is equal. This can happen if the MDP is layered and the number of parents is the same for all nodes in all layers. Thus, the maximum entropy backward subsumes the uniform backward.
\end{remark}

 Note that the uniform backward is not always maximum entropy. For instance, as shown in Appendix~\ref{appen:n}, it is not for any of the MDPs we analyze.

One major difference between GSQL and maximum entropy GFNs is what happens if $n$ is not learned to high fidelity. If $n(s)$ is greater than zero, then maximum entropy GFNs will still be GFNs yet GSQL does not have this property.

Lastly, unlike the uniform backward policy, the maximum entropy backward relaxes the reachability assumption. This is because $n$ is zero for unreachable states. However, this may require sampling backward trajectories similar to \cite{zhang2022generative}, which is beyond the scope of this work.

\subsection{Remarks on PCL}

\cite{nachum2017bridging} showed that if (\ref{eq:pcl}) holds for all sub-trajectories of a certain length, then the soft Bellman equation holds and vice versa. We extend the result to full trajectories.

\begin{propositionrep}
    If PCL holds for all sub-trajectories reaching a terminal state ($\traj[i][T]$), all trajectories starting at an initial state ($\traj[0][i]$), or all full trajectories ($\traj[0][T]$), then it holds for all sub-trajectories. We call these conditions the terminal, initial, and trajectory PCL conditions, respectively.
\end{propositionrep}
\begin{proof}
    If PCL holds for two trajectories $\traj[i][T]$ and $\traj[i+1][T]$, then it holds for $s_{i}a_is_{i+1}$, thus if terminal PCL holds, PCL holds as the soft Bellman equation holds. By symmetry, the same is true for initial PCL.

    Assuming trajectory PCL, for a trajectory $\traj[0][T]$, we define $V(s_i)$ as
    \begin{equation}\label{eq:tpcl}
    V(s_0)+\sum_{t=0}^{i-1} \gamma^{t} \left[\tau \log \pi(a_t|s_t) - R(s_t, a_t)\right].
    \end{equation}
    We first show that this definition is consistent in the sense that if there is another trajectory $\traj[0][k][s'][a']a'_ks_i$, then $V(s_i)$, defined in (\ref{eq:tpcl}), is also equal to 
    \begin{equation}
V(s'_0)+\sum_{t=0}^{k} \gamma^{t} \left[\tau \log \pi(a'_t|s'_t) - R(s'_t, a'_t)\right].
    \end{equation}
    
    Since trajectory PCL holds, and
    \begin{equation}\label{eq:somef234}
    V(s'_0)+\sum_{t=0}^{k} \gamma^{t} \left[\tau \log \pi(a'_t|s'_t) - R(s'_t, a'_t)\right] + \sum_{t=i}^{T-1} \gamma^{t} \left[\tau \log \pi(a_t|s_t) - R(s_t, a_t)\right]
    \end{equation} 
    and
    \begin{equation}\label{eq:somef2334}
        V(s_0)+\sum_{t=0}^{T-1} \gamma^{t} \left[\tau \log \pi(a_t|s_t) - R(s_t, a_t)\right]
    \end{equation}
    are equal to $V(T)$ subtracting them yields zero. By subtracting (\ref{eq:somef2334}) and (\ref{eq:somef234}) and removing the duplicate part, we are left with the two alternative definitions of $V(s_i)$, which means they must be equal.

    Since all PCL holds for all initial trajectories, then PCL holds.
\end{proof}

We can use PCL to calculate $n$ using Proposition~\ref{prop:nrevsql}. 

\begin{propositionrep}
Given $n$, maximum entropy GFNs with (\ref{eq:trajectorybalance}) and GSQL with trajectory \ref{eq:pcl} have the same residual and gradient.
\end{propositionrep}
\begin{proof}
    The backward probability of any trajectory $\traj$ is $1/n(s_T)$ by Remark~\ref{remark:nsub}. Thus maxent GFNs have the objective
    $\log Z + \sum_{i=0}^{T-1} \log \pi(a_t|s_t) = \unnormprob(s_T) - \log n(s_T)$ which is (\ref{eq:pcl}) for GSQL .
\end{proof}

\begin{figure*}
\begin{subfigure}[T]{\textwidth}
\centering
\includegraphics[width=0.5\textwidth]{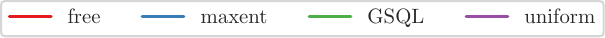}
\end{subfigure}
\begin{subfigure}[T]{0.33\textwidth}
\centering
\includegraphics[width=\textwidth]{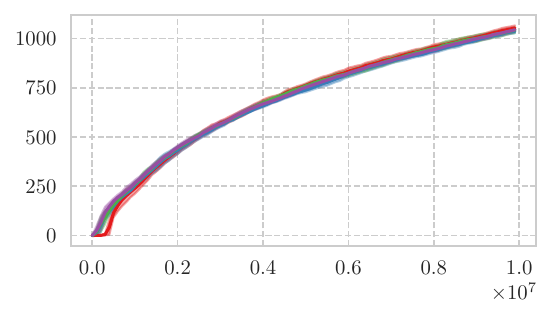}
\end{subfigure}
\begin{subfigure}[T]{0.33\textwidth}
\centering
\includegraphics[width=\textwidth]{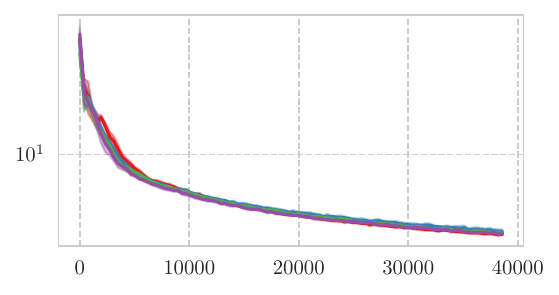}
\end{subfigure}
\begin{subfigure}[T]{0.33\textwidth}
\centering
\includegraphics[width=\textwidth]{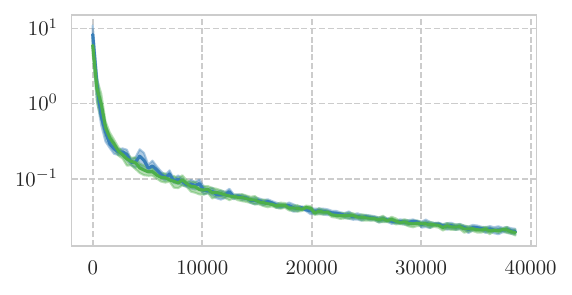}
\end{subfigure}
\begin{subfigure}[T]{0.33\textwidth}
\centering
\includegraphics[width=\textwidth]{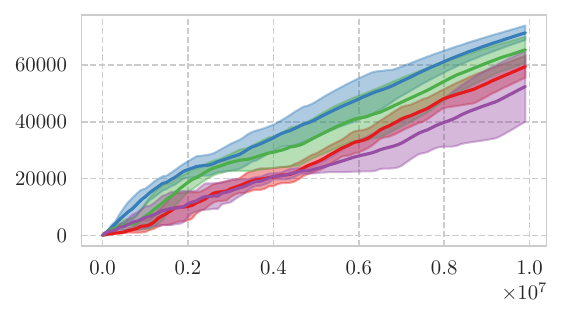}
\caption{number of modes found\label{fig:modes}}
\end{subfigure}
\begin{subfigure}[T]{0.33\textwidth}
\centering
\includegraphics[width=\textwidth]{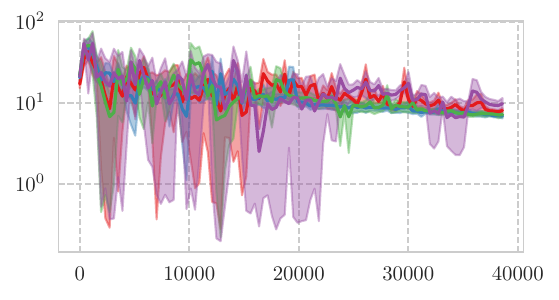}
\caption{policy loss \label{fig:pi_loss}}
\end{subfigure}
\begin{subfigure}[T]{0.33\textwidth}
\centering
\includegraphics[width=\textwidth]{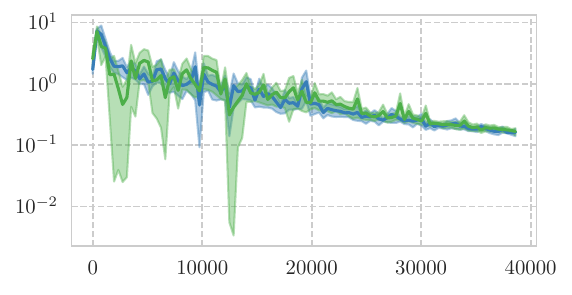}
\caption{$n$ loss \label{fig:n_loss}}
\end{subfigure}

\caption{Experiment statistics. Confidence intervals show the IQM. From top to bottom, the rows belong to the sEH and QM9 experiments.}
\end{figure*}

\section{EXPERIMENTS}

This section is divided into three subsections, each focusing on MDPs that do not follow the assumptions of \cite{zhang2022generative}. They help illustrate that it is not hard to find MDPs where the uniform backward is not maximum entropy. We compare maximum entropy GFNs and GSQL with GFNs with uniform and learned backward policies. 

Appendix~\ref{appen:exp} presents all experiment details, pseudocode, and further experiments.

\subsection{A simple MDP}

In the MDP presented in Figure~\ref{fig:unfi}, there are three paths from the initial state $s_0$ to the destination $s_T$: $s_0,s_1,s_T$, $s_0,s_2,s_1,s_T$, and $s_0,s_2,s_T$. The uniform backward gives $q(s_1|s_T)=q(s_2|s_T)=q(s_0|s_1)=q(s_2|s_1)=1/2$; thus, the paths have probabilities 0.25, 0.25, and 0.5, respectively, yielding an entropy of $3/2\log 2$. The maximum entropy GFN yields the backward 
$q(s_1|s_T)=2/3$ and $q(s_2|s_T)=1/3$ since there are 2 paths from $s_1$ but only one path from $s_2$. Furthermore, $q(s_2|s_1)=q(s_0|s_1)=1/2$, since only one path per parent exists. The entropy is $\log 3$, which is the maximum entropy.

\subsection{Hypergrid}
We now focus on the hypergrid domain from \cite{bengio2021flow}. The state is a vector that starts at $0$. At each step, the agent can increase one of the components of the vector by one, i.e., move one step in one of the directions or terminate the episode. The agent is in a bounded box, and the target function is given by 
\begin{equation}\label{eq:rew_hg}
\resizebox{.9 \columnwidth}{!}{$
    0.1 + 
    0.5 \prod_i \mb I[0.25 < |s_i - 0.5|] + 
    2 \prod_i \mb I[0.3 < |s_i - 0.5| < 0.4],
    $
}
\end{equation}
for the indicator function $\mb I$ and the ratio of the current position and the maximum allowed position in component $i$, $s_i$. In Figure~\ref{fig:64}, for a $64\times64$ grid, we show the target, the logarithm of the number of paths $l$, and the marginal distribution of uniform backward policy and the maximum entropy GFNs. First, note that $l$ gets very big. We cannot store $n=\exp(l)$ as a 64-bit integer as $n$ for the top right corner is $(2\times63)!/63!/63!$. Second, notice how the maximum entropy GFN distributes the flow more evenly and does not accumulate flow on the bottom and left edges.

We test the models on the $64^2$ and $8^4$ hypergrids to reach terminal states in proportion to (\ref{eq:rew_hg}). The results can be seen in Figure \ref{fig:enter-label2}. In the $64^2$ experiments, GSQL failed to reach all modes but worked fine on the $8^4$ grid, whereas maximum entropy GFN successfully learns to sample in proportion to the target. The failure of GSQL is minor as it can be alleviated with added exploration, yet it illustrates the difference between GSQL and maximum entropy GFNs. Indeed, maximum entropy GFNs, unlike GSQL, are GFNs regardless of the quality of the estimated $n$. We note that since GSQL fails to learn the target distribution in the $64^2$ experiment, the fact that its entropy is higher than the maximum entropy is irrelevant.

\begin{table*}    
\caption{KL divergence of the policy of different models on small sEH task of \cite{shen2023towards}. The cells show the average KL divergence between the row and column policies.}
    \label{tab:seh18:kl}
    \centering
    \resizebox{\textwidth}{!}
    {%
\begin{tabular}{ll
S[table-format=2.2(2)]
S[table-format=2.2(2)]
S[table-format=2.2(2)]
S[table-format=2.2(2)]
S[table-format=2.2(2)]
S[table-format=2.2(2)]}
\toprule
                       &            & {free}   & {uniform}    & \multicolumn{2}{c}{maxent}  & \multicolumn{2}{c}{GSQL}     \\
                       &          $n$    &   {}     & {}   & {known} & {learned} &  {known} & {learned }\\
\midrule
free                   &            &  0.32\pm0.06 & 12.61\pm0.23& 11.03\pm0.23 & 11.27\pm0.23 & 11.13\pm0.23 & 11.29\pm0.22  \\
uniform                &            & 10.69\pm0.37& \bfseries 0.08\pm0.01 &  0.38\pm0.01 &  0.41\pm0.01 &  0.39\pm0.01 &  0.40\pm0.02  \\
\midrule
\multirow[t]{2}{*}{maxent} & known    & 10.02\pm0.37 &  0.45\pm0.01 & \bfseries 0.05\pm0.01 &  \bfseries 0.09\pm0.01 & \bfseries 0.08\pm0.01 &   \bfseries 0.10\pm0.01  \\
                       & learned  &  9.83\pm0.36&  0.53\pm0.03  & \bfseries 0.09\pm0.01 & \bfseries 0.05\pm0.01 &   \bfseries0.10\pm0.01 & \bfseries 0.07\pm0.01 \\
\midrule
\multirow[t]{2}{*}{GSQL}   & known   & 10.22\pm0.37 &  0.56\pm0.02 & \bfseries 0.10\pm0.01 & \bfseries 0.12\pm0.01 & \bfseries 0.09\pm0.01 & \bfseries 0.12\pm0.01 \\
                       & learned &  9.94\pm0.37 &  0.52\pm0.01 &  \bfseries 0.11\pm0.01 & \bfseries 0.09\pm0.01 &  \bfseries0.11\pm0.01 & \bfseries 0.08\pm0.01 \\
\bottomrule
\end{tabular}   }

\end{table*}

\subsection{Molecule design}

In this section, we examine the sEH task of \citet{bengio2021flow} and the QM9 task of \citet{jain2023multi}.  We use a tree-building MDP in the sEH \citep{jin2018junction} and a graph-building MDP in the QM9 experiments. In the tree-building environment, every state is a tree, and each action either adds a node and connects it to an existing node or sets an attribute on each edge. In the sEH experiments, each node represents a fragment (collection of atoms), and edge features show the place of connection of two fragments. In the graph-building environment, every state is a connected graph (every node is reachable from every other node), and actions either add a new node and edge or set the attribute on edge.

The graph-building environment is more expressive than the tree-building environment. However, unlike the tree-building environment, the graph-building environment can lead to molecules that cannot exist and are invalid.

The tree-building MDP is much more structured, and we can calculate $n$ for each state directly to verify that we are, in fact, learning (see Appendix~\ref{appen:n}). Figure~\ref{fig:n_vs_baseline} shows the mean squared error between the predicted $\log n$ and the real $\log n$. The difference between the learned $n$ and the ground truth is small.

\begin{figure}
    \centering
    \includegraphics[width=\columnwidth]{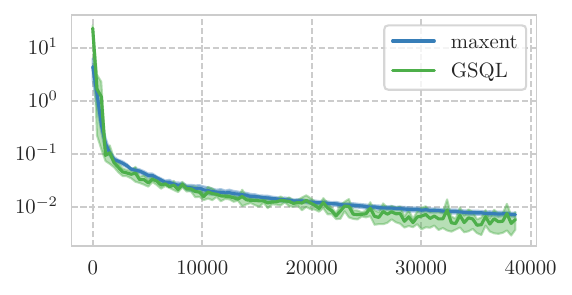}
    \caption{MSE of the learned $n$ and the ground truth in the sEH experiments. 
    \label{fig:n_vs_baseline}}
\end{figure}

We show the KL divergence between the policies found using different methods in Table~\ref{tab:seh18:kl} when trained using the small sEH MDP of \cite{shen2023towards}. Notice how GFNs with free backward policy have a higher KL divergence among themselves than the other policies. Indeed, this results from the GFN constraints having infinite feasible solutions for this MDP. As expected, GSQL and maximum entropy GFNs find close policies. We also note that they are close regardless of whether $n$ is given or learned. Lastly, as shown by the higher KL divergence between the GFNs with uniform backward policy and maximum entropy GFNs, the other two find different policies. 

Next, we focus on the full sEH task; we show the GFN and $n$ losses in the top row of Figure~\ref{fig:pi_loss} and Figure~\ref{fig:n_loss}. Not only $n$ is learnable, but it is also is easier to learn than the policy constraint as the reward is always zero. As shown in the top row of Figure~\ref{fig:modes}, GFNs with free backward policy start finding modes with a short lag. We show some statistics after training in Table~\ref{tab:seh:train}. The top $K$ statistics are the same for all policies thus we can infer that all models are GFNs regardless of the backward but not using a free backward policy helps at the initial phase of the training and using maximum entropy backward policy policy increases the entropy. We refer to Appendix~\ref{appen:exp} for a definition of the top $K$ statistics.

Our last set of experiments focuses on the QM9 experiments. While we calculated $n$ for the sEH experiments, we do not do so here as it is prohibitively expensive to do via the general recursion, and we could not find a combinatorial structure to exploit. As shown in the second row of Figure~\ref{fig:pi_loss} and Figure~\ref{fig:n_loss}, it is harder than the sEH experiment, but the $n$ objective is still easier than the policy objective. The bottom row of Figure~\ref{fig:modes} shows that maxent GFNs outperform GSQL, and GSQL outperforms the other two GFNs. Table~\ref{tab:qm9:train} shows some statistics. We note that we put a hard limit of 100,000 modes as the cost of finding modes grows quadratically. In this experiment, maxent GFN and GSQL backward find more modes. Again, the key takeaway is that as we deviate more from the assumptions of \citet{zhang2022generative}, the performance of GFNs with uniform backward worsen.

\begin{table*} 
    
\centering
\caption{sEH experiment results.\label{tab:seh:train}}
    \resizebox{\textwidth}{!}{%
\begin{tabular}{ll
S[table-format=2.2(2)]
S[table-format=1.2(2)]
S[table-format=1.2(2)]
S[table-format=1.2(2)]
S[table-format=1.2(2)]
S[table-format=3.2(2)]
S[table-format=3.2(2)]
}
\toprule
 model & $n$ & {entropy} & {diverse top $K$ reward} & {top $K$ diversity} & {top $K$ reward}  & {modes with $\unnormprob\geq1$}  & {modes with $\unnormprob\geq0.875$}  \\
\midrule
free & none & 77.92\pm0.02 & 1.03\pm0.00 & 0.46\pm0.00 & 1.03\pm0.00 & \bfseries1049.90\pm5.66 & 3378.00\pm24.30 \\
uniform & none & 77.27\pm0.02 & 1.03\pm0.00 & 0.46\pm0.00 & 1.03\pm0.00 & 1042.90\pm5.21 & 3318.30\pm31.15 \\
\cmidrule{1-8}
\multirow[c]{2}{*}{GSQL} & learned & \bfseries78.42\pm0.02 & 1.03\pm0.00 & 0.46\pm0.00 & 1.03\pm0.00 & 1043.90\pm7.18 & \bfseries3446.50\pm40.27 \\
 & known & \bfseries78.39\pm0.01 & 1.03\pm0.00 & 0.46\pm0.00 & 1.03\pm0.00 & 1036.60\pm7.74 & 3379.90\pm29.55 \\
\cmidrule{1-8}
\multirow[c]{2}{*}{maxent} & learned & \bfseries78.41\pm0.03 & 1.03\pm0.00 & 0.46\pm0.00 & 1.03\pm0.00 & 1035.70\pm5.56 & 3417.60\pm32.82 \\
 & known & \bfseries78.41\pm0.02 & 1.03\pm0.00 & 0.46\pm0.00 & 1.03\pm0.00 & 1042.50\pm5.98 & \bfseries3344.00\pm44.67 \\
\bottomrule
\end{tabular}}

\end{table*}
\raggedbottom
\begin{table*} 
\caption{QM9 experiment results.\label{tab:qm9:train}}
    \resizebox{\textwidth}{!}{%
\begin{tabular}{l
S[table-format=2.2(2)]
S[table-format=1.2(2)]
S[table-format=1.2(2)]
S[table-format=1.2(2)]
S[table-format=5(4)]
S[table-format=6]
}
\toprule
 model & {entropy} & {diverse top $K$ reward} & {top $K$ diversity} & {top k reward}  & {modes with $\unnormprob\geq1.125$}  & {modes with $\unnormprob\geq1$}  \\
\midrule
free & 93.41\pm0.24 & \bfseries1.20\pm0.00 & 0.83\pm0.00 & \bfseries1.20\pm0.00 & 60465\pm2157 & 100000\pm0 \\
uniform & 89.96\pm7.61 & 1.19\pm0.01 & 0.81\pm0.04 & 1.19\pm0.01 & 52865\pm4481 & 100000\pm0 \\
GSQL & \bfseries98.59\pm0.30 & 1.19\pm0.00 & \bfseries0.84\pm0.00 & 1.19\pm0.00 & 63909\pm2730 & 100000\pm0 \\
maxent & \bfseries98.22\pm0.11 & \bfseries1.20\pm0.00 & 0.83\pm0.00 & \bfseries1.20\pm0.00 & \bfseries 71339\pm1468 & 100000\pm0 \\
\bottomrule
\end{tabular}}

\end{table*}

\section{CONCLUSION}

This work showed an equivalence between entropy-regularized RL and GFNs. Concretely, we showed that using the number of trajectories to each terminal state, we can create a reward such that the probability of reaching terminal states of entropy-regularized RL is proportional to an unnormalized distribution. We use the equivalence to show the equivalence of (\ref{eq:trajectorybalance}) and (\ref{eq:pcl}) for a specific class GFNs and entropy regularized RL with our proposed reward.

Building on top of our proposed extension of entropy regularized RL, we introduced maximum entropy GFNs, subsuming the uniform backward policy of \citet{zhang2022generative}. A benefit of our model is that it automatically falls back to the uniform backward policy whenever the uniform backward policy is maximum entropy. We showed that we can learn $n$ using the soft Bellman equation on the inverted MDP, as calculating it via the recursion may be too expensive, and a combinatorial structure may be hard to identify, as is the case with the graph-building environment used in the QM9 experiments. We then verified that $n$ is indeed learnable empirically. 

While we mirror the assumptions of \citet{bengio2021flow}, our analysis is limited to undiscounted MDPs with deterministic transitions and only one initial state. Solving the same problem with stochastic transitions may be of interest \citep[e.g.][]{pan2023stochastic} but is more complex. For instance, while the reachability of the terminal states guarantees a solution in the deterministic case, a solution may not exist in the stochastic formulation of \citet{pan2023stochastic}. \citet{jiralerspong2023expected}'s formulation for GFNs in stochastic MDPs overcome the feasibility issue of \citet{pan2023stochastic}'s formulation and maximum entropy GFNs may help relax the tree assumption of \citet{jiralerspong2023expected} yet maintain their equilibrium results. However, it is unclear how to define the corresponding inverse MDP properly.

Independently, \citet{tiapkin2023generative} and \citet{deleu2024discrete} provide an alternative way of obtaining GFNs from soft  Q-learning given a fixed backward policy. These methods can be used in conjunction with the maximum entropy backward policy introduced here. Limiting ourselves to maximum entropy makes our proposed method simpler and does not conceptually depend on GFNs. With the equivalence results, many RL algorithms like Munchausen \citet{vieillard2020munchausen} are now directly applicable to GFNs. Future work will look at the improvements those methods can bring. 

\citet{derman2021twice} pointed out that entropy-regularized RL is equivalent to a particular robust reinforcement learning formulation. However, it is unclear how such a result would apply meaningfully here as the uncertainty set that \citet{derman2021twice} proposes would give non-zero reward to transitions, something we cannot have with GFNs.

Current GFN environments are too structured. Most non-toy environments have many of the features \citet{zhang2022generative} require. For instance, the MDPs we analyze are all layered but do not have the same number of parents. In the sEH experiment, the bulk of the high-reward molecules are in the final layer. We expect as the MDPs used become more and more complex, maximum entropy GFNs shine more as they did in QM9 experiment compared to the sEH experiment.

\ackaccepted{We thank Nikolay Malkin for comments about uniform backward policy and Gabriele Farina for comments about dilated entropy. This research was supported by compute resources provided by  Calcul Quebec (\url{calculquebec.ca}), the BC DRI Group, the Digital Research Alliance of Canada (\url{alliancecan.ca}), and Mila (\url{mila.quebec}).}
\bibliographystyle{apalike}
\bibliography{ref}

\clearpage

 \begin{enumerate}

 \item For all models and algorithms presented, check if you include:
 \begin{enumerate}
   \item A clear description of the mathematical setting, assumptions, algorithm, and/or model. Yes
   \item An analysis of the properties and complexity (time, space, sample size) of any algorithm. No
   \item (Optional) Anonymized source code, with specification of all dependencies, including external libraries. Yes
 \end{enumerate}

 \item For any theoretical claim, check if you include:
 \begin{enumerate}
   \item Statements of the full set of assumptions of all theoretical results. Yes
   \item Complete proofs of all theoretical results. Yes
   \item Clear explanations of any assumptions. Yes
 \end{enumerate}

 \item For all figures and tables that present empirical results, check if you include:
 \begin{enumerate}
   \item The code, data, and instructions needed to reproduce the main experimental results (either in the supplemental material or as a URL). Yes
   \item All the training details (e.g., data splits, hyperparameters, how they were chosen). No
         \item A clear definition of the specific measure or statistics and error bars (e.g., with respect to the random seed after running experiments multiple times). Yes
         \item A description of the computing infrastructure used. (e.g., type of GPUs, internal cluster, or cloud provider). No
 \end{enumerate}

 \item If you are using existing assets (e.g., code, data, models) or curating/releasing new assets, check if you include:
 \begin{enumerate}
   \item Citations of the creator If your work uses existing assets. Not Applicable
   \item The license information of the assets, if applicable. Not Applicable
   \item New assets either in the supplemental material or as a URL, if applicable. Yes
   \item Information about consent from data providers/curators. Not Applicable
   \item Discussion of sensible content if applicable, e.g., personally identifiable information or offensive content. Not Applicable
 \end{enumerate}

 \item If you used crowdsourcing or conducted research with human subjects, check if you include:
 \begin{enumerate}
   \item The full text of instructions given to participants and screenshots. Not Applicable
   \item Descriptions of potential participant risks, with links to Institutional Review Board (IRB) approvals if applicable. Not Applicable
   \item The estimated hourly wage paid to participants and the total amount spent on participant compensation. Not Applicable
 \end{enumerate}

 \end{enumerate}

\nosectionappendix
\begin{toappendix}

\section{GENERATIVE FLOW NETWORKS}\label{appen:gfn}

    \begin{table}
    \caption{Notation \label{tab:notation}}
    \centering
    \begin{tabular}{ll}
    \toprule
        Unnomralized distribution on terminal states & $\unnormprob$ \\
        Normalizing factor & $Z$ or $F(s_0)$ \\
        Normalized version of $\unnormprob$ & $\normprob$ \\
        Probability simplex over the set $\mc X$ & $\Delta(\mc X)$ \\
        Calligraphic letters are used for sets & $\mc X, \mc S, \mc A$ \\
        Set of states & $\mc S$ \\
        Set of initial states & $\mc S_0 \subset \mc S$ \\
        Initial state (if unique, i.e., $\mc S_0 =\{s_0\}$) & $s_0$ \\
        Set of actions & $\mc A$ \\
        Function that returns a set & $\md A, \md L$ \\
        Action mask function & $\md A$ \\
        Set of terminal states & $\term \subset \mc S$ \\
        Non terminal states & $\bar{\term}$ \\
        Power set & $P$ \\
        Deterministic transition function & $\trans$ \\
        Initial state distribution & $\mb P_0$ \\
        Set of trajectories leading to $s$ from a unique starting state $s_0$ & $\md L(s)$ \\
        Number of trajectories leading to $s$ from a unique starting state $s_0$ & $n(s)$ \\
        $\log n(s)$ & $l(s)$  \\
        Marginal distribution & $\mu$ \\
        Discount factor & $\gamma$ \\

        \bottomrule
    \end{tabular}
\end{table}

\begin{table}
\caption{Abvreviations\label{tab:abrev}}
    \centering
    \begin{tabular}{ll}
    \toprule
        Markov decision process & MDP \\
        Generalized value function & GVF \\
        Generative flow networks & GFN \\
        Soft Q-learning & SQL \\
        Flow matching & FM \\
        Detailed balance & DB \\
        Trajectory balance & TB \\
        Path consistency learning & PCL \\
        \bottomrule
    \end{tabular}
    
\end{table}

In this section, we review a few concepts related to GFN that did not fit in the main text due to to lack of space. While they are not necessary to understand this work, we belive that they can help understand GFNs better.

Linearizing the term $F(s)\pi(a|s)$ of (\ref{eq:detailedbalance}) yields the flow action function $F(s,a)$ that has to fit the flow matching constraints \citep{bengio2021flow}
\begin{equation}\label{eq:flowmatching}\tag{FM}
    \unnormprob(s) + \sum_{\mathclap{a\in\md A(s)}} F(s, a) = \sum_{\mathclap{s',a' \in \paren(s)}} F(s',a'),
\end{equation}
for each state $s$ where $F:\mc S \times \mc A \rightarrow \mb{R}_{\geq0}$ is the state action flow function and $\unnormprob$ is zero at nonterminal states. Any flow function $F$, the policy $\pi$, and the backward policy $q$ that fit in (\ref{eq:detailedbalance}) or (\ref{eq:flowmatching}) fit in $F(s,a) =F(s)\pi(a|s)$, $F(s,a)=F(s')q(s,a|s')$, $F(t)= \unnormprob(t)$ and $F(s_0) = Z$ for a transition triplet $s'=T(s,a)$ and terminal state $t$. 

Sub-trajectory balance (STB) \citep{madan2023learning} is the convex combination of (\ref{eq:trajectorybalance}) over all sub-trajectories. The sub-TB constraint is defined as 
\begin{equation} \label{eq:subtb}\tag{STB}
    F(s_i)\prod_{t=i}^{j-1} \pi(a_t|s_t) = F(s_j) \prod_{t=i}^{j-1} q(s_i,a_i|s_{t+1}),
\end{equation}
for all partial trajectory $\traj[i][j]$. When learning with (\ref{eq:subtb}), \citep{madan2023learning} propose taking the weighted sum of the residual of all equations. They propose weighting the samples by $\gamma$ to the length of the sub-trajectories as it corresponds to the expected loss of a random sub-trajectory length that is sampled from a geometric distribution with parameter $\gamma$ \citep{shwartz2001death}. In this work, we assume $\gamma=1$.

The constraints (\ref{eq:flowmatching}), (\ref{eq:detailedbalance}), (\ref{eq:subtb}), and (\ref{eq:trajectorybalance}) are equivalent in the sense if one of them holds, all of them hold \citep{bengio2021gflownet,malkin2022trajectory,madan2023learning}. 

\paragraph{Multiple solutions for GFNs.} Any convex combination of forward policies of GFNs leads to another forward policy of a GFN. Thus, if there are two valid solutions, there are infinite solutions. If two trajectories lead to the same node, then there are infinite solutions as the flow of that node can either flow in one or the other trajectory. An example is the MDP shown in Figure~\ref{fig:unfi}.

\section{CALCULATING \texorpdfstring{$n(s)$} FOR MDPS WITH COMBINATORIAL STRUCTURES}\label{appen:n}

Calculating $n(s)$ may be possible using the combinatorial structure of the state space. We give a small list in Table~\ref{tab:n}. The DAG MDP is based on the work of \cite{deleu2022bayesian}. The other MDPs are inspired by \citet{bengio2021flow}. These formulas are  sensitive to the definition of the problem; for instance, calculating $n$ for DAGs, where the actions add edges and nodes, is more complicated than where the nodes are fixed.

Similarly, assuming that the graph stays connected makes calculating $n$ more complicated. For instance, we are unaware of any reasonable way that would not require traversing the MDP to calculate $n$ for general connected graph-building environments. This is one of the reasons why we advocate learning $n$ instead of calculating it.

\begin{table}
\caption{$n$ for MDPs with combinatorial structure. Bars indicate that the variable is fixed.}
\label{tab:n}
    \centering
    \begin{tabular}{llc}
    \toprule
State space & Action space & $n(s)$ \\
\midrule
Words & append from right & $1$  \\
Words of length $N$ & append from either sides & $2^{N-1}$  \\
DAGs with $\bar{N}$ nodes and $E$ edges & connect two nodes & $E!$ \\
Trees & add node & See (\ref{eq:trees}) \\
\bottomrule
\end{tabular}

\end{table}

The number of ways to define a tree represented as a directed graph rooted at a node $r$ fits the following DP
\begin{equation}\label{eq:trees0}
    W_r[s] = \frac{(D_r[s] - 1)!}{
    \prod_{c \in \text{Children}(s)} D_r[c]!}\prod_{c \in \text{Children}(s)} W_r[c],
\end{equation}
where $D_r[c]$ is the number of children of node $c$ and fits $D_r[s]=1 + \sum_{c\in \text{Children}} D_r[c]$. We obtain
(\ref{eq:trees0}) by shuffling all actions that can be used to build the children (it holds since the actions used to build the children are independent in a tree). The total number of trajectories is
\begin{equation}\label{eq:trees}
    \sum_{s \in\text{Nodes}} W_s[s].
\end{equation}

We note that if attributes are on the edges, $D_r[c]$ is not the number of children but the number of children and attributes.

Lastly, we show in Figure~\ref{fig:uniform_not_maxent} that for tree-building environments, similar to the drug design problem of \cite{bengio2021flow}, the uniform backward is not maximum entropy. The proposed state has two non-isomorphic parents whose $n$ differ.

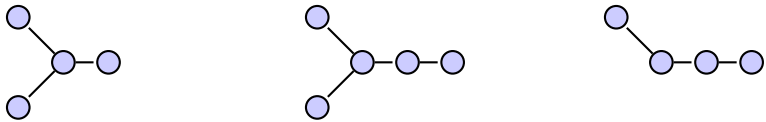
\begin{figure}
    \centering
    \begin{subfigure}[T]{0.225\textwidth}
	\begin{tikzpicture}[-,>=stealth',shorten >=1pt,auto,node distance=3cm,
		thick,main node/.style={circle,fill=blue!20,draw,minimum size=3mm,inner sep=0pt},scale=0.6]
		 \node[main node] (1) at (0,0) {};
		\node[main node] (2) at (-1,1) {};
		\node[main node] (3) at (-1,-1) {};
		\node[main node] (4) at (1,0) {};
		
		\path[every node/.style={font=\sffamily\small}]
		(1) edge node {} (2)
		(1) edge node {} (3)
		(1) edge node {} (4);
	\end{tikzpicture}
 \end{subfigure}
     \begin{subfigure}[T]{0.225\textwidth}
	\begin{tikzpicture}[-,>=stealth',shorten >=1pt,auto,node distance=3cm,
		thick,main node/.style={circle,fill=blue!20,draw,minimum size=3mm,inner sep=0pt},scale=0.6]
		 \node[main node] (1) at (0,0) {};
		\node[main node] (2) at (-1,1) {};
		\node[main node] (3) at (-1,-1) {};
		\node[main node] (4) at (1,0) {};
        \node[main node] (5) at (2,0) {};
		
		\path[every node/.style={font=\sffamily\small}]
		(1) edge node {} (2)
		(1) edge node {} (3)
		(1) edge node {} (4)
		(4) edge node {} (5);
	\end{tikzpicture}
 \end{subfigure}
     \begin{subfigure}[T]{0.225\textwidth}
	\begin{tikzpicture}[-,>=stealth',shorten >=1pt,auto,node distance=3cm,
		thick,main node/.style={circle,fill=blue!20,draw,minimum size=3mm,inner sep=0pt},scale=0.6]
		 \node[main node] (1) at (0,0) {};
		\node[main node] (2) at (-1,1) {};
		\node[main node] (4) at (1,0) {};
        \node[main node] (5) at (2,0) {};
		
		\path[every node/.style={font=\sffamily\small}]
		(1) edge node {} (2)
		(1) edge node {} (4)
		(4) edge node {} (5);
	\end{tikzpicture}
 \end{subfigure}
    \caption{Proof that the uniform backward is not maximum entropy on tree-building environments. Left and right are the parents of the middle tree. Assuming nodes are unique, the left tree has 12 trajectories that reach it, while the right tree has 8.\label{fig:uniform_not_maxent}}
\end{figure}

\section{EXPERIMENT DETAILS \label{appen:exp}}

The sEH experiments are tree-building environments where we assume that each state is a tree, each node in the tree is a fragment, a group of atoms, and each edge has a feature to describe how two fragments are connected \citep{jin2018junction}. The fragments are those of \citet{bengio2021flow} or \citet{shen2023towards}. The reward is \citet{bengio2021flow}'s proxy for predicting the binding energy of a molecule to soluble epoxide hydrolase (sEH). We refer to \citet{bengio2021flow} for a more detailed explanation.

The QM9 experiments are graph-building environments where we assume that each state is a connected graph (i.e., there is a path between each pair of nodes). Each node represents a carbon, nitrogen, fluorine, or oxygen atom. The main difference with the sEH experiments is that the state graph can have loops and terminal states may be invalid molecules. The target is \cite{jain2023multi}'s proxy, a neural network predicting the HUMO-LUMO gap trained on the QM9 dataset \citep{ramakrishnan2014quantum}.

We transform the rewards into a scalar where zero is the worst quantity, and the reward is below one for most molecules. We divide the output of the proxy of \citet{bengio2021flow} by 8. We linearly rescale the QM9 gaps such that the 5\% molecules with the lowest gap in \citet{ramakrishnan2014quantum}'s dataset have a reward greater than one while the rest have a reward between 0 and 1. For the QM9, since the MDP can create invalid molecules, we give a reward of $\exp(-75)$ to invalid molecules.

\paragraph{Metrics.}
For the top $K$, diverse top $K$, and top $K$ diversity, we sample $N$ terminal states. We then calculate the average top $K$ rewards, the average top $K$ rewards under the constraint that the molecules have a similarity of less than 0.5 and one minus the average similarity of the top $K$ molecules. If no $K$ valid solutions exist, we assume the worst (i.e., the reward is zero or the molecule is one of the existing molecules for diversity). For molecules, we use the Tanimoto similarity of RDkit \citep{landrum2013rdkit}.

\paragraph{Calculating (\ref{eq:subtb}) and (\ref{eq:pcl}).} We use dynamic programming to calculate these objectives in a performant manner. Algorithm~\ref{algo:pcl} shows an implementation using functions available in PyTorch \citep{paszke2019pytorch}. The main idea is that we can calculate the matrix $D_{ij}=\sum_{t=i}^{j}x_t$ by subtracting the vector $y_i=\sum_{t=0}^ix_t$ from its transposed shifted. Here $v$ is the value and $x_i$ is $\log \pi(a_i|s_i) - \log q(s_i,a_i|s_{i+1})$ for GFNs and $\log \pi(a_i|s_i)$ for SQL. We can avoid reversing the trajectory when learning $n$ by using $x_i= -\log\pi(s_i, a_i|s_{i+1})$.

\begin{algorithm}
\caption{Psudocode for fast (\ref{eq:subtb}) and (\ref{eq:pcl})\label{algo:pcl}}
\begin{minted}{python}
def subtb(v, x):
    return torch.triu(v[:-1, None] - v[None, 1:] + cross(x))
def cross(x):
    y = torch.cumsum(x, 0)
    return y[None] - shift_right(y)[:, None]
def shift_right(x): 
    x = torch.roll(x, 1, dims=0)
    x[0] = 0
    return x
\end{minted}
\end{algorithm}

\paragraph{A practical algorithm} A training loop roughly looks like the following
\begin{enumerate}
    \item Obtain a batch of trajectories. The batch can either be sampled directly or through a more complex sampling scheme like the one proposed by \cite{kim2023local}. The batch can also be obtained by obtaining a set of terminal states (potentially from a replay buffer) and using the backward policy to sample a backward trajectory like \citet{zhang2022generative}.
    \item Minimize the residual of a constraint on $n$ like PCL or the soft Bellman equation.
    \item Minimize the residual of a GFN or GSQL constraint using PCL, TB, DB, or any other variant.
    \item Update the sampling model.
\end{enumerate}
For speed, we do steps 2 and 3 simultaneously by having a single neural network with two heads. We use the exponential moving average of the weights of the model as the sampling model

\paragraph{On the choice of the hyperparameters.} 

We opt for (\ref{eq:trajectorybalance}) and trajectory (\ref{eq:pcl}) to learn $n$ and the policy. Since we use the Huber loss function \citep{huber1992robust} defined as
\begin{equation}
    \text{Huber}(x;\delta) = \begin{cases}
    0.5x^2 / \delta & \text{if } |x| \leq \beta \\
    \beta (|x| - 0.5\beta)/\delta & \text{otherwise}  ,
    \end{cases}
\end{equation}

we add both losses together and do not need to worry about one loss contributing too much more to the loss. The main benefit of using the Huber loss is that the contribution of objectives to the gradient is more balanced than the mean squared error if the magnitude of the objectives is different, as the gradients are bounded. Using the Huber loss allows us to reduce the gradient clipping needed for stable training and improve the performance of all GFNs. We use the same hyperparameters for all models. Table~\ref{tab:hpsl} shows the hyperparameters used.

\begin{table}
\caption{Hyperparameters used}
\label{tab:hpsl}
\centering
\begin{tabular}{ll}
\toprule
Learning Rate & $5e-4$ \\
Reward Exponent & 96 \\
Batch Size & 256 \\
$\epsilon$ uniform (exploration) & $1e-3$\\
Samples & $1e7$ \\
EMA sampling model & $0.95$ \\
Huber loss $\delta$ & $0.25$ \\
Huber loss $\beta$ & $1$ \\
\bottomrule
\end{tabular}
\end{table}

\paragraph{Code and reproducibility.}

The code is available at \url{https://github.com/recursionpharma/gflownet}

\paragraph{Quality of the solutions.}
Table~\ref{tab:pearnsonr} shows the Pearson correlation coefficient of different backward policies. We limit the models to 5 fragments from the 18 fragments of \cite{shen2023towards} in the sEH experiment to be able to calculate the total probability of any terminal state. We sample molecules directly from $\unnormprob$ and use them to calculate the Pearson correlation coefficient. It is worth noting how GFNs with free backward have a lower correlation coefficient than the rest when used with (\ref{eq:subtb}). 
\begin{table}
\centering
\caption{Pearson correlation coefficient of different backward policies. We sample terminal states with a random walk and then either sample them proportionately to the objective (proportional) or uniformly (uniform). $ST$ and $T$ mean sub-trajectory and trajectory level constraint.}
\label{tab:pearnsonr}
\begin{tabular}{llll
S[table-format=1.2(2)]
S[table-format=1.2(2)]
}
\toprule
model & $n$ & loss & $\epsilon$ random & {Person (proportional)} & {Person (uniform)}\\
\midrule
\multirow[t]{4}{*}{free}  
 &  & \multirow[t]{2}{*}{ST} 
       & 0.001 & 0.82\pm0.01 & 0.88\pm0.01 \\
 &  &  & 0.01  & 0.82\pm0.01 & 0.87\pm0.00 \\
 &  &  \multirow[t]{2}{*}{T}
       & 0.001 & 0.92\pm0.00 & 0.92\pm0.00 \\
 &  &  & 0.01  & 0.92\pm0.00 & 0.91\pm0.00 \\
\midrule
\multirow[t]{4}{*}{uniform}
 &  & \multirow[t]{2}{*}{ST}
       & 0.001 & 0.89\pm0.00 & 0.89\pm0.00\\
 &  &  & 0.01  & 0.90\pm0.00 & 0.90\pm0.00\\
 &  &  \multirow[t]{2}{*}{T}
       & 0.001 & 0.92\pm0.00 & 0.92\pm0.00 \\
 &  &  & 0.01  & 0.93\pm0.00 & 0.91\pm0.00\\
\midrule
\multirow[t]{8}{*}{maxent}
 & \multirow[t]{4}{*}{known} & \multirow[t]{2}{*}{ST}
       & 0.001 & 0.89\pm0.00 & 0.90\pm0.00 \\
 &  &  & 0.01  & 0.89\pm0.00 & 0.90\pm0.00 \\
 &  & \multirow[t]{2}{*}{T}
       & 0.001 & 0.93\pm0.00 & 0.92\pm0.00\\
 &  &  & 0.01  & 0.93\pm0.00 & 0.92\pm0.00\\
 \cmidrule(lr){2-6}
 & \multirow[t]{4}{*}{learned} & \multirow[t]{2}{*}{ST}
       & 0.001 & 0.86\pm0.00 & 0.88\pm0.00 \\
 &  &  & 0.01  & 0.85\pm0.01 & 0.87\pm0.00\\
 &  & \multirow[t]{2}{*}{T}
       & 0.001 & 0.92\pm0.00 & 0.91\pm0.00 \\
 &  &  & 0.01  & 0.92\pm0.00 & 0.92\pm0.00\\
 \midrule
\multirow[t]{8}{*}{GSQL}
 & \multirow[t]{4}{*}{known} & \multirow[t]{2}{*}{ST}
       & 0.001 & 0.88\pm0.00 & 0.89\pm0.00 \\
 &  &  & 0.01  & 0.87\pm0.00 & 0.88\pm0.00\\
 &  & \multirow[t]{2}{*}{T}
       & 0.001 & 0.93\pm0.00 & 0.92\pm0.00\\
 &  &  & 0.01  & 0.93\pm0.00 & 0.91\pm0.00\\
 \cmidrule(lr){2-6}
 & \multirow[t]{4}{*}{learned} & \multirow[t]{2}{*}{ST}
       & 0.001 & 0.83\pm0.00 & 0.86\pm0.00\\
 &  &  & 0.01  & 0.83\pm0.01 & 0.86\pm0.00\\
 &  & \multirow[t]{2}{*}{T}
       & 0.001 & 0.93\pm0.00 & 0.92\\
 &  &  & 0.01  & 0.93\pm0.00 & 0.91\pm0.00 \\
\bottomrule
\end{tabular}
\end{table}

Table~\ref{tab:seh_full} and Table~\ref{tab:qm9:full} provide statistics for different parameters. These table have \textbf{not} been used for hyper parameter tuning.
\begin{table}
\caption{Statistics for the sEH experiment.}
    \label{tab:seh_full}
    \centering
    \resizebox{\textwidth}{!}{%
    \begin{tabular}{llll
S[table-format=1.2(2)]S[table-format=1.2(2)]
S[table-format=1.2(2)]S[table-format=1.2(2)]
S[table-format=1.2(2)]S[table-format=1.2(2)]
}
\toprule
 model & $\pi$ loss & $n$ loss & $\epsilon$ random & {entropy} & {diverse top $K$ reward} & {top $K$ diversity} & {top $K$ reward}  & {modes with $\unnormprob\geq1$}  & {modes with $\unnormprob\geq.875$}  \\
 \midrule
\multirow[c]{4}{*}{free} & \multirow[c]{2}{*}{ST} & \multirow[c]{2}{*}{none} & 0.001 & 72.39\pm0.75 & 1.03\pm0.00 & 0.44\pm0.00 & 1.03\pm0.00 & 646.20\pm11.29 & 4007.80\pm31.68 \\
 &  &  & 0.01 & 68.22\pm0.61 & 1.02\pm0.00 & 0.44\pm0.00 & 1.02\pm0.00 & 565.00\pm10.19 & 4114.20\pm45.65 \\
\cmidrule{2-10}
 & \multirow[c]{2}{*}{T} & \multirow[c]{2}{*}{none} & 0.001 & 77.92\pm0.02 & 1.03\pm0.00 & 0.46\pm0.00 & 1.03\pm0.00 & 1049.90\pm5.66 & 3378.00\pm24.30 \\
 &  &  & 0.01 & 77.61\pm0.06 & 1.03\pm0.00 & 0.46\pm0.00 & 1.03\pm0.00 & 941.30\pm6.43 & 3382.20\pm55.95 \\
\cmidrule{1-10}
\multirow[c]{4}{*}{uniform} & \multirow[c]{2}{*}{ST} & \multirow[c]{2}{*}{none} & 0.001 & 77.22\pm0.03 & 1.03\pm0.00 & 0.44\pm0.00 & 1.03\pm0.00 & 668.80\pm12.22 & 4059.40\pm35.78 \\
 &  &  & 0.01 & 77.22\pm0.02 & 1.03\pm0.00 & 0.44\pm0.00 & 1.03\pm0.00 & 649.11\pm9.53 & 4316.33\pm50.56 \\
\cmidrule{2-10}
 & \multirow[c]{2}{*}{T} & \multirow[c]{2}{*}{none} & 0.001 & 77.27\pm0.02 & 1.03\pm0.00 & 0.46\pm0.00 & 1.03\pm0.00 & 1042.90\pm5.21 & 3318.30\pm31.15 \\
 &  &  & 0.01 & 77.24\pm0.02 & 1.03\pm0.00 & 0.46\pm0.00 & 1.03\pm0.00 & 971.10\pm7.23 & 3324.70\pm25.68 \\
\cmidrule{1-10}
\multirow[c]{10}{*}{GSQL} & \multirow[c]{4}{*}{ST} & \multirow[c]{2}{*}{ST} & 0.001 & 78.59\pm0.05 & 1.02\pm0.00 & 0.43\pm0.00 & 1.02\pm0.00 & 488.50\pm7.00 & 4076.70\pm30.72 \\
 &  &  & 0.01 & 78.60\pm0.07 & 1.02\pm0.00 & 0.43\pm0.00 & 1.02\pm0.00 & 507.50\pm17.67 & 4305.50\pm58.98 \\
\cmidrule{3-10}
 &  & \multirow[c]{2}{*}{known} & 0.001 & 78.46\pm0.02 & 1.03\pm0.00 & 0.44\pm0.00 & 1.03\pm0.00 & 565.11\pm10.36 & 4093.89\pm26.39 \\
 &  &  & 0.01 & 78.55\pm0.03 & 1.02\pm0.00 & 0.43\pm0.00 & 1.02\pm0.00 & 486.50\pm14.49 & 4151.70\pm72.83 \\
\cmidrule{2-10}
 & \multirow[c]{6}{*}{T} & \multirow[c]{2}{*}{ST} & 0.001 & 78.37\pm0.01 & 1.03\pm0.00 & 0.46\pm0.00 & 1.03\pm0.00 & 1053.60\pm4.74 & 3489.00\pm30.93 \\
 &  &  & 0.01 & 78.38\pm0.02 & 1.03\pm0.00 & 0.46\pm0.00 & 1.03\pm0.00 & 972.00\pm6.65 & 3388.90\pm20.39 \\
\cmidrule{3-10}
 &  & \multirow[c]{2}{*}{T} & 0.001 & 78.42\pm0.02 & 1.03\pm0.00 & 0.46\pm0.00 & 1.03\pm0.00 & 1043.90\pm7.18 & 3446.50\pm40.27 \\
 &  &  & 0.01 & 78.38\pm0.02 & 1.03\pm0.00 & 0.46\pm0.00 & 1.03\pm0.00 & 958.40\pm7.94 & 3389.60\pm46.82 \\
\cmidrule{3-10}
 &  & \multirow[c]{2}{*}{known} & 0.001 & 78.39\pm0.01 & 1.03\pm0.00 & 0.46\pm0.00 & 1.03\pm0.00 & 1036.60\pm7.74 & 3379.90\pm29.55 \\
 &  &  & 0.01 & 78.37\pm0.02 & 1.03\pm0.00 & 0.46\pm0.00 & 1.03\pm0.00 & 931.11\pm6.92 & 3293.44\pm24.53 \\
\cmidrule{1-10}
\multirow[c]{10}{*}{maxent} & \multirow[c]{4}{*}{ST} & \multirow[c]{2}{*}{ST} & 0.001 & 78.41\pm0.02 & 1.03\pm0.00 & 0.44\pm0.00 & 1.03\pm0.00 & 554.70\pm9.22 & 4134.80\pm20.68 \\
 &  &  & 0.01 & 78.48\pm0.03 & 1.02\pm0.00 & 0.43\pm0.00 & 1.02\pm0.00 & 549.50\pm7.45 & 4302.20\pm34.57 \\
\cmidrule{3-10}
 &  & \multirow[c]{2}{*}{known} & 0.001 & 78.41\pm0.02 & 1.03\pm0.00 & 0.44\pm0.00 & 1.03\pm0.00 & 656.50\pm12.83 & 3993.30\pm25.19 \\
 &  &  & 0.01 & 78.47\pm0.02 & 1.03\pm0.00 & 0.44\pm0.00 & 1.03\pm0.00 & 609.70\pm9.72 & 4153.50\pm35.74 \\
\cmidrule{2-10}
 & \multirow[c]{6}{*}{T} & \multirow[c]{2}{*}{ST} & 0.001 & 78.42\pm0.01 & 1.03\pm0.00 & 0.46\pm0.00 & 1.03\pm0.00 & 1048.50\pm5.80 & 3388.60\pm40.80 \\
 &  &  & 0.01 & 78.40\pm0.02 & 1.03\pm0.00 & 0.46\pm0.00 & 1.03\pm0.00 & 963.90\pm6.12 & 3373.70\pm26.76 \\
\cmidrule{3-10}
 &  & \multirow[c]{2}{*}{T} & 0.001 & 78.41\pm0.03 & 1.03\pm0.00 & 0.46\pm0.00 & 1.03\pm0.00 & 1035.70\pm5.56 & 3417.60\pm32.82 \\
 &  &  & 0.01 & 78.40\pm0.01 & 1.03\pm0.00 & 0.46\pm0.00 & 1.03\pm0.00 & 951.60\pm7.66 & 3354.00\pm19.02 \\
\cmidrule{3-10}
 &  & \multirow[c]{2}{*}{known} & 0.001 & 78.41\pm0.02 & 1.03\pm0.00 & 0.46\pm0.00 & 1.03\pm0.00 & 1042.50\pm5.98 & 3344.00\pm44.67 \\
 &  &  & 0.01 & 78.41\pm0.02 & 1.03\pm0.00 & 0.46\pm0.00 & 1.03\pm0.00 & 949.78\pm6.22 & 3296.00\pm35.13 \\
\bottomrule
\end{tabular}%
}
    
\end{table}

\begin{table}[]
    \centering
    \caption{Statistics for the QM9 experiment.}\label{tab:qm9:full}
    \resizebox{\textwidth}{!}{%
    \begin{tabular}{llll
S[table-format=1.2(2)]S[table-format=1.2(2)]
S[table-format=1.2(2)]S[table-format=1.2(2)]
S[table-format=5.2(5)]S[table-format=5.2(4)]
}
\toprule
model & $\pi$ loss & $n$ loss & $\epsilon$ random & {entropy} & {diverse top $K$ reward} & {top $K$ diversity} & {top $K$ reward}  & {modes with $\unnormprob\geq1.125$}  & {modes with $\unnormprob\geq1$}\\
\midrule
\multirow[c]{4}{*}{free} & \multirow[c]{2}{*}{ST} & \multirow[c]{2}{*}{none} & 0.001 & 89.72\pm8.93 & 1.15\pm0.01 & 0.80\pm0.08 & 1.15\pm0.01 & 36941.80\pm2828.57 & 100000.00\pm0.00 \\
 &  &  & 0.01 & 97.78\pm0.41 & 1.13\pm0.00 & 0.89\pm0.00 & 1.13\pm0.00 & 39792.20\pm1304.05 & 100000.00\pm0.00 \\
\cmidrule{2-10}
 & \multirow[c]{2}{*}{T} & \multirow[c]{2}{*}{none} & 0.001 & 93.41\pm0.24 & 1.20\pm0.00 & 0.83\pm0.00 & 1.20\pm0.00 & 60465.30\pm2157.86 & 100000.00\pm0.00 \\
 &  &  & 0.01 & 83.35\pm8.04 & 1.21\pm0.01 & 0.75\pm0.08 & 1.21\pm0.01 & 62732.10\pm1845.51 & 100000.00\pm0.00 \\
\cmidrule{1-10}
\multirow[c]{4}{*}{uniform} & \multirow[c]{2}{*}{ST} & \multirow[c]{2}{*}{none} & 0.001 & 96.82\pm4.13 & 1.14\pm0.01 & 0.88\pm0.01 & 1.14\pm0.01 & 45371.30\pm2548.19 & 100000.00\pm0.00 \\
 &  &  & 0.01 & 100.91\pm0.04 & 1.14\pm0.00 & 0.89\pm0.00 & 1.14\pm0.00 & 46464.80\pm615.89 & 100000.00\pm0.00 \\
\cmidrule{2-10}
 & \multirow[c]{2}{*}{T} & \multirow[c]{2}{*}{none} & 0.001 & 89.96\pm7.61 & 1.19\pm0.01 & 0.81\pm0.04 & 1.19\pm0.01 & 52865.10\pm4481.11 & 100000.00\pm0.00 \\
 &  &  & 0.01 & 96.89\pm0.05 & 1.20\pm0.00 & 0.82\pm0.00 & 1.20\pm0.00 & 70884.20\pm824.47 & 100000.00\pm0.00 \\
\cmidrule{1-10}
\multirow[c]{6}{*}{GSQL} & \multirow[c]{2}{*}{ST} & \multirow[c]{2}{*}{ST} & 0.001 & 102.58\pm0.10 & 1.14\pm0.00 & 0.88\pm0.00 & 1.14\pm0.00 & 46815.80\pm1470.56 & 100000.00\pm0.00 \\
 &  &  & 0.01 & 102.67\pm0.05 & 1.14\pm0.00 & 0.88\pm0.00 & 1.14\pm0.00 & 45283.30\pm861.96 & 100000.00\pm0.00 \\
\cmidrule{2-10}
 & \multirow[c]{4}{*}{T} & \multirow[c]{2}{*}{ST} & 0.001 & 100.29\pm2.39 & 1.07\pm0.11 & 0.85\pm0.01 & 1.07\pm0.11 & 56199.90\pm3737.65 & 100000.00\pm0.00 \\
 &  &  & 0.01 & 99.12\pm0.29 & 1.19\pm0.00 & 0.83\pm0.01 & 1.19\pm0.00 & 54856.40\pm1866.60 & 100000.00\pm0.00 \\
\cmidrule{3-10}
 &  & \multirow[c]{2}{*}{T} & 0.001 & 98.59\pm0.30 & 1.19\pm0.00 & 0.84\pm0.00 & 1.19\pm0.00 & 63909.30\pm2730.65 & 100000.00\pm0.00 \\
 &  &  & 0.01 & 89.18\pm8.78 & 1.21\pm0.01 & 0.74\pm0.08 & 1.21\pm0.01 & 67928.20\pm2021.95 & 100000.00\pm0.00 \\
\cmidrule{1-10}
\multirow[c]{6}{*}{maxent} & \multirow[c]{2}{*}{ST} & \multirow[c]{2}{*}{ST} & 0.001 & 102.61\pm0.04 & 1.14\pm0.00 & 0.88\pm0.00 & 1.14\pm0.00 & 47538.90\pm1432.97 & 100000.00\pm0.00 \\
 &  &  & 0.01 & 102.48\pm0.04 & 1.14\pm0.00 & 0.88\pm0.00 & 1.14\pm0.00 & 46710.10\pm482.77 & 100000.00\pm0.00 \\
\cmidrule{2-10}
 & \multirow[c]{4}{*}{T} & \multirow[c]{2}{*}{ST} & 0.001 & 98.49\pm0.30 & 1.20\pm0.00 & 0.83\pm0.01 & 1.20\pm0.00 & 67809.20\pm2652.35 & 100000.00\pm0.00 \\
 &  &  & 0.01 & 98.12\pm0.09 & 1.20\pm0.00 & 0.82\pm0.00 & 1.20\pm0.00 & 71374.50\pm1106.31 & 100000.00\pm0.00 \\
\cmidrule{3-10}
 &  & \multirow[c]{2}{*}{T} & 0.001 & 98.22\pm0.11 & 1.20\pm0.00 & 0.83\pm0.00 & 1.20\pm0.00 & 71339.40\pm1468.82 & 100000.00\pm0.00 \\
 &  &  & 0.01 & 98.31\pm0.11 & 1.20\pm0.00 & 0.83\pm0.00 & 1.20\pm0.00 & 71962.60\pm1064.17 & 100000.00\pm0.00 \\ 
\bottomrule
\end{tabular}%
}
\end{table}

\end{toappendix}

\end{document}